\documentclass[12pt]{iopart}

\usepackage{dsfont}
\usepackage[utf8]{inputenc}
\usepackage[english]{babel}
\usepackage{mathtools}
\usepackage{nicefrac}
\usepackage{csquotes}
\usepackage{amssymb,amsthm}
\usepackage{mathrsfs,amsfonts}
\usepackage{array}
\usepackage{listings}
\usepackage{float}
\usepackage{algorithm2e}
\usepackage{hyperref}
\usepackage{caption}

\def\code#1{\texttt{#1}}

\usepackage[normalem]{ulem}

\newtheorem{theorem}{Theorem}
\newtheorem{definition}{Definition}

\usepackage{subfig}
\usepackage{graphicx}
\ifpdf
  \DeclareGraphicsExtensions{.pdf,.pdf,.png,.jpg}
\else
  \DeclareGraphicsExtensions{.pdf}
\fi

\usepackage{xcolor}

\begin{document}

\title[Efficient quantitative assessment of robot swarms]{Efficient quantitative assessment of robot swarms: coverage and targeting L\'{e}vy strategies}

\author{S~Duncan{\footnote[1]{These authors contributed equally to this article.}}$^1$, {G~\mbox{Estrada--Rodriguez\dag}}$^2$, J~Stocek\dag$^3$, M~Dragone$^1$, P~A~Vargas$^1$ and {H~Gimperlein}$^4$}

\address{$^1$ Robotics Lab, Edinburgh Centre for Robotics, School of Mathematical and Computer Sciences \& School of Engineering and Physical Sciences, Heriot--Watt University, Edinburgh, EH14 4AS, UK
}
\address{$^2$ Basque Center for Applied Mathematics, Alameda de Mazarredo 14, 48009 Bilbao, Bizkaia, Spain
}
\address{$^3$ British Antarctic Survey, High Cross, Madingley Road, Cambridge, CB3 0ET, UK
}
\address{$^4$ Engineering Mathematics, Leopold-Franzens-Universit\"{a}t Innsbruck, Technikerstra\ss e 13, 6020 Innsbruck, Austria
}
\eads{\mailto{sd246@hw.ac.uk}, \mailto{gestrada@bcamath.org}, \mailto{jaksto@bas.ac.uk}, \mailto{m.dragone@hw.ac.uk}, \mailto{p.a.vargas@hw.ac.uk}, \mailto{heiko.gimperlein@uibk.ac.at}}

\begin{abstract}
Biologically inspired \emph{strategies} have long been adapted to swarm robotic systems, including biased random walks, reaction to chemotactic cues and long-range coordination. In this paper we \textcolor{black}{apply}  \emph{analysis tools} developed for  modeling biological systems, such as continuum descriptions, \textcolor{black}{to} the efficient quantitative characterization of robot swarms. As an illustration, \textcolor{black}{both Brownian} and L\'{e}vy strategies with a characteristic long-range movement are discussed. As a result we obtain computationally fast methods for the optimization of robot movement laws to achieve a prescribed collective behavior. We show how to compute performance metrics like coverage and hitting times, and illustrate the accuracy and efficiency of our approach for area coverage and search problems. Comparisons between the continuum model and robotic simulations confirm the quantitative agreement and speed up of our approach. Results confirm and quantify the advantage of L\'{e}vy strategies over Brownian motion for search and area coverage problems in swarm robotics.
\end{abstract}
\\
\maketitle
\section{Introduction}

{S}{earch} and coverage tasks are the basis for many potential applications for biologically-inspired robot swarms, for instance in search and rescue \cite{kantor2003distributed}, foraging for natural resources \cite{liu2007modeling}, exploration/mapping \cite{marjovi2009multi}, environmental monitoring \cite{schmickl2011cocoro}, cooperative cleaning \cite{altshuler2005swarm}, and intruder detection and surveillance scenarios \cite{kolias2011swarm}.

Multiple robots can spread across an environment in order to sense it from different vantage points and distribute themselves to maximize the rate at which the environment is explored. Like their biological models, robots in these swarms are governed by simple reactive behaviours, usually by availing only of local and instantaneous information and requiring limited or no communication with each other \cite{fredslund2001robot}. Since robots do not need to build shared models of their environment, nor globally agreed plans, target systems can be more robust and fault tolerant, and potentially more scalable and more suitable to unknown and dynamic environments than centralised coordination approaches \cite{parker2008distributed}.

However, all this flexibility comes at a price. Crucially, the biggest hurdle for successfully applying swarm robotic approaches in general is the inherent difficulty in engineering the behaviour of each robot to achieve system-level properties \cite{brambilla}. Behavior-based design is the most common approach to develop a swarm robotics system \cite{parker1995design}. Control laws for individual robots’ behaviours are implemented and successively improved \textcolor{black}{sometimes by a process of trial and error \cite{francesca2016automatic}}, or by relying on soft-computing methods such as multi-robot reinforcement-learning \cite{mataric1997reinforcement} and evolutionary algorithms used in evolutionary robotics \cite{nolfi2000evolutionary,vargas2014her}, until the overall swarm exhibits the desired global behaviour. 
Simulations are usually necessary, as these methods require many iterations to converge and it would be simply \textcolor{black}{im}practical and take an extremely long time to run numerous field tests with many real robots. Consequently, the development of robotic swarms is usually limited to specific scenarios where both tasks and environment are strictly defined. Unfortunately, although simulations are much faster than real experiments, the greater the number of robots to be simulated, the longer it takes to obtain results, \textcolor{black}{especially at higher densities or for longer-range interactions.} \textcolor{black}{Even if simulators like ARGoS manage to simulate thousands of robots in specific situations \cite{Argos}, e.g.~through parallel execution of subspaces, their run time will scale with the number of robots.} Hence, it is hard to provide and predict the effect of changes either in each robot's behaviour, or in the number of robots in the overall swarm performance. 

\textcolor{black}{Math-based approaches} such as the one presented in this paper address these problems by \textcolor{black}{combining} fundamental  principles  of modeling  and  control  for  large scale robotic swarms,  and providing  analytical tools  that can be used to accelerate the design of robot swarms behaviours \cite{milutinovi2006modeling}. \textcolor{black}{While in \cite{estrada2018swarming} \iffalse we gave \fi a detailed derivation of the mathematical model for a swarm robotic system {was given}, a comparison with robotic simulations was not considered.}

Specifically, in the work presented here we tackle these problems using tools developed for the modeling of biological systems. We start from the movement of the individual robot, including local communication and interactions. As an example, we derive a continuum model for  L\'{e}vy \cite{DBLP:journals/corr/KrivonosovDZ16} strategies with characteristic long-range movements\textcolor{black}{, between the ballistic (L\'{e}vy exponent $\alpha=1$) and diffusive ($\alpha=2$) regimes.} {The resulting continuum model describes the evolution of the probability density of robots in terms of a non-local macroscopic diffusion equation. It allows efficient quantitative characterization of robot swarms and scalability for large number of robots since we are not studying the behaviour of each individual but rather the collective motion of a robot's density. Macroscopic \textcolor{black}{partial differential equations (PDEs)} not only provide fast computational approaches, but also provide analytical insights into the system-level behaviour of the robot collective \cite{lerman2004review}.}


{The main results in this paper are the following:
\begin{itemize}
    \item \textcolor{black}{Robotic quantities of interest, such as area coverage and the expected hitting time to reach a target, can be computed from the macroscopic robot density, as illustrated in Section 6.}
    
    \item The \textcolor{black}{numerical experiments} illustrate quantitative agreement for these quantities with individual robot simulations within the statistical margins of error.
    \item From the numerical experiments we confirm the advantages of strategies characterized by a strong presence of long trajectories (L\'{e}vy walks) over Brownian motion. \textcolor{black}{For the strategies considered in this article, the L\'{e}vy exponent $\alpha =1$ is observed to lead to optimal area coverage and hitting times.}
    \item Our numerical implementation provides a fast method to predict the global performance of a swarm robotic system.
\end{itemize}}

\textcolor{black}{In particular, our analytical models and computational tools allow to efficiently assess the consequences of swarm robotic design decisions with a complexity independent of the number of robots, as long as their density is low. This paper thereby shows the insights and methods which can be obtained by adapting ideas from continuum models of realistic interacting particle systems to robotic problems. Detailed models for concrete applications will be pursued elsewhere.}

The remainder of the paper is organised as follows: Section II provides an overview of related work. Section III introduces the search strategy, while the derivation of the macroscopic description is the content of Section IV. Section V describes the robotic simulations and Section VI the comparison with the numerical results. In Section VII we discuss the results and Section VII summarises our conclusions.

\section{Related work}

Simple approaches such as bacterial chemotaxis, phototaxis and run-and-tumble processes have been applied to individual robots \cite{nurzaman2009yuragi}. For searching large areas for sparse and randomly located targets, a combination of biased random walk and L\'{e}vy strategies has proven to be efficient and robust to changes in the environment, for instance, in the case of underwater multi-robot systems \cite{sutantyo2013collective}.

From a different perspective, in swarm robotic systems L\'{e}vy strategies have also proven to be effective and provide a controlled model system to understand biological behavior. These strategies involve long-range movement, mimicking  the behavior of  T cells \cite{fricke2016immune}. They prove useful for target search problems for sparsely distributed targets. Further examples where  bio-inspired search strategies have been applied to robots include \cite{turduev2010chemical,dhariwal2004bacterium}.


For biological systems like flocks of birds or fish schools, self-organization and pattern formation have motivated the development of mathematical models to reproduce and analyze the collective behavior.  An influential mathematical model governing these interacting particle systems was proposed by T. Vicsek et al.~\cite{vicsek1995novel} in 1995. In this model, particles move with a constant speed and at each time step they align their velocity to the average velocity of the neighboring particles with some random perturbation.  Subsequent works like \cite{couzin2002collective} and \cite{degond2008continuum} added realistic refinements  to the mathematical description. 

Second order models, where the velocity changes dynamically depending on the interactions and alignment, have been studied in great length \cite{cucker2007emergent,chuang2007state,ha2008particle}, where also relevant macroscopic systems of equations have been derived.

\textcolor{black}{
Applications of these ideas in the context of robotic systems are starting to emerge. 
In \cite{milutinovi2006modeling} 
a proof of principle of the use of macroscopic modeling for the control of a large system of robots was demonstrated. The main focus was not on modeling details of the robotic system, but on its optimal control by a centralized controller. Cues to guide the movement of the robots were included in \cite{Hamann2006analytical} to direct the movement of robots. As these first works modeled the actual movement strategies and interactions only in a generic way, discrepancies between the macroscopic equation and individual particle models were observed.\\
More detailed models for the aggregation of robots were proposed in \cite{soysal2006macroscopic}, while neglecting the modeling of the spatial movement. Similarly, \cite{prorok2011multi} started from a detailed non-spatial model for a search problem, to which diffusive and drift components where added. 
}

\textcolor{black}{
The first macroscopic models more closely related to our work have only emerged recently \cite{andersonquantitative,elamvazhuthi2018pde,elamvazhuthi2018nonlinear,zhang2018performance}. The focus in these works is towards realistic metrics to measure the performance of a swarm robotic system, as a basis for its optimization and control. Non-interacting Brownian robots are studied as the simplest possible model systems, with a first extension to nonlinear diffusion in \cite{elamvazhuthi2018nonlinear}. Simple kinetic models for chemotaxis related to the current work have been studied in \cite{mes1,mes}. The approach was used for the design of strategies for assembling the robots according to a given target probability density. We refer to \cite{Elamvazhuthi_2019} for the opportunities of partial differential equation models in robotics. }\textcolor{black}{On the other hand, there has been some works on L\'{e}vy search strategies, for example see \cite{elamvazhuthi2016coverage,fioriti2015levy,marthaler2004levy}.}

\textcolor{black}{Our current work \textcolor{black}{considers} the modeling  of realistic, complex robot movement by macroscopic partial differential equations on large spatial and time scales, as relevant for the design of interactions and movement laws. We illustrate the modeling opportunities for the performance metrics of coverage and hitting times. 
}

\section{Description of the search strategy}\label{sec: movement strategy}

A swarm robotics system consists of a large number of simple independent robots with local rules, communication and interactions among them and with the environment, where the local interactions may lead to collective behaviour of the swarm. In this work, we use a swarm of N \textit{e-puck} robots \cite{epuck}, \textcolor{black}{which are differential drive robots}, in a domain $\Omega$ in $\mathbb{R}^2$ with reflective boundary conditions, which provides a specific model system.  Each robot of diameter $\varrho>0$ is characterized by its position $\mathbf{x}_i\in\mathds{R}^2$, its run time $\tau_i$, \textcolor{black}{which is defined as the time between two consecutive changes in direction}, and direction $\mathbf{\theta}_i\in S=\{|\mathbf{x}_i|=1 \}\subseteq\mathds{R}^2$. We assume that each robots moves according to the following rules:
\begin{enumerate}
\item[A1.]  The trajectories are characterized by long straight line motion interrupted by instantaneous reorientations \textcolor{black}{when the robot stops (for a very short time)}, and here a new direction of motion is randomly chosen. This movement is called a velocity jump process. The speed $c$ of the robots is assumed to be constant.
\item[A2.] Starting at position $\mathbf{x}_i$ at time $t$, a robot $i$ runs in
direction $\theta_i$ for a L\'{e}vy distributed time $\tau_i$, called the \enquote{run time}.
\item[A3.] The run time $\tau_i$ follows a power-law \textcolor{black}{survival function with cumulative density}
\begin{equation}
\psi_i(\mathbf{x}_i,\tau_i)=\left(\frac{\varsigma_0(\mathbf{x}_i)}{\varsigma_0(\mathbf{x}_i)+\tau_i}\right)^{\alpha},\ \alpha\in(1,2)\ . \label{eq: survival}
\end{equation}
It describes the probability that an individual moving in some fixed direction does not stop until time $\tau_i$. \textcolor{black}{The function $\varsigma_0(\mathbf{x}_i)>0$ captures possible spatial inhomogeneities.}
\item[A4.] After being running for a time $\tau_i$, the robots stop (see A1.) with a frequency given by 
\begin{equation}
    \beta_i(\mathbf{x}_i,\tau_i)=-\frac{\partial_{\tau_i}\psi_i}{\psi_i}=\frac{\varphi_i}{\psi_i}\ ,\label{eq: beta}
\end{equation}
\textcolor{black}{where $\varphi_i$ is the stopping density function.}
\item[A5.] Each time the robot stops it selects a new direction $\theta_i^*$ according to 
a \textcolor{black}{symmetric} distribution $k(\mathbf{\theta}_i;\mathbf{\theta}_i^*)=\tilde{k}(|\theta_i^*-\theta_i|)$. This describes a bias according to the previous orientation which incorporates an element of persistence of orientation.

\item[A6.] The collisions of two robots are assumed to be elastic. The new direction after collision is $\theta'_i=\theta_i-2(\theta_i\cdot\nu)\nu$, where $\nu=\frac{\mathbf{x}_i-\mathbf{x}_j}{|\mathbf{x}_i-\mathbf{x}_j|}$ is the normal vector at the point of collision.

\end{enumerate}

The power law behaviour in Assumption 3 describes the long tailed distribution of run times, instead of the Poisson process in classical velocity jump processes \cite{othmer2000diffusion}.

Note that the assumptions correspond to independent individuals with simple capabilities relative to typical tasks for swarm robotic systems. In \textcolor{black}{the systems considered in this article} robots interact only with their neighbors in a narrow sensing region, \textcolor{black}{determined by the range of the physical sensors}. The movement decisions are based on the current positions and velocities, not information from earlier interactions.  This assures the scalability to large numbers of robots, while non-local collective movement may emerge from local rules \cite{senanayake2016search}.

\textcolor{black}{While the short ranged interactions correspond to the specific systems described in Sections 5 and 6, where the sensor range is around $6$cm and of the order of the robot's diameter, macroscopic models can be derived also for systems with long-range interactions \cite{estrada2018swarming}.}

Related movement laws have been used for target search, for example, in the experiments in \cite{fricke2016immune}.  Refined local control laws and the possibility for quantitative experiments with robots open up novel modeling opportunities. In Section \ref{sec: comparison} we apply the theoretical results and present numerical experiments. 

{In this article \emph{microscopic} quantities (and equations) are used to refer to individual robots (and the trajectories of the individual robots), while \emph{macroscopic} quantities (and equations) describe the {density} of robots (and the time evolution of the density).}

\section{Microscopic description for individual  movement}\label{sec: collision description}
In this section we adapt the mathematical description developed in \cite{estrada2018swarming} to the specific experimental set up.

For the  $N$-individual system described in Section \ref{sec: movement strategy}, the microscopic density $\sigma=\sigma(\mathbf{x}_i,t,\theta_i,\tau_i)$ of robots in position $\mathbf{x}_i$ at time $t$ moving in the direction $\theta_i$ for some time $\tau_i$, evolves according to a kinetic equation 
\begin{equation}
    \partial_t\sigma+\sum_{i=1}^N\left(\partial_{\tau_i}+c\theta_i\cdot\nabla_{\mathbf{x}_i}\right)\sigma=-\sum_{i=1}^N\beta_i\sigma\ ,\label{eq: N particles}
\end{equation}
in the domain $\Omega^N = \{(\mathbf{x}_1, ..., \mathbf{x}_N) \in \mathds{R}^{2\times N}:\ |\mathbf{x}_i-\mathbf{x}_j|\geq\varrho \ \forall i,j \}$. The left hand side of (\ref{eq: N particles}) describes the trajectories followed by the robots, more specifically, a straight line motion. The right hand side gives the density of robots that stop with a frequency $\beta_i$, given in (\ref{eq: beta}).

After stopping, according to Assumption 5~robots choose a new direction of motion according to the turn angle operator $T_i$ given by
\begin{equation}
    T_i\phi(\theta_i^*)=\int_S k(\mathbf{\theta}_i;\mathbf{\theta}_i^*)\phi(\theta_i)d\theta_i \label{eq: turn angle operator}\ ,
\end{equation}
where the new direction $\theta_i^*$ is symmetrically distributed with respect to the previous
direction $\theta_i$, according to the distribution $k(\mathbf{\theta}_i;\mathbf{\theta}_i^*)=\tilde{k}(|\theta_i^*-\theta_i|)$ \textcolor{black}{and $\phi(\theta_i)\in L^2(S)$}  \cite{alt1980biased}. The expression (\ref{eq: turn angle operator}) describes the change in direction $\theta_i\to\theta_i^*$. See Appendix \ref{sec: turn_angle_properties} for some properties of the operator $T$.

\subsection{Equation for the two-particle density}\label{sec: two particle case}

The description (\ref{eq: N particles}) of the $N$-robot problem \emph{a priori} requires the understanding of collisions among the whole system. In this section, however, we aim for a macroscopic description for low densities where collisions of more than two individuals may be neglected \cite{cercignani2013mathematical}. Hence, by neglecting collisions of $3$ or more robots we integrate out robots $3,...,N$ from $\sigma$.  The transport equation which describes the movement of two individuals {$\mathbf{x}_1,\ \mathbf{x}_2\in \Omega^2$}, is given by 
\begin{align}
    \partial_{\tau_1}\sigma+\partial_{\tau_2}\sigma+\partial_t\sigma+c\theta_1\cdot\nabla_{\mathbf{x}_1}\sigma &+c\theta_2\cdot\nabla_{\mathbf{x}_2}\sigma\nonumber\\&=-(\beta_1+\beta_2)\sigma \label{eq: initial model}\ .
\end{align}
Here $\sigma=\sigma(\mathbf{x}_1,\mathbf{x}_2,t,\theta_1,\theta_2,\tau_1,\tau_2)$ is the two-particle density function.

We also define a density independent of $\tau_2$ given by 
\[\tilde{\sigma}_{\tau_1}(\mathbf{x}_1,\mathbf{x}_2,t,\theta_1,\theta_2,\tau_1)=\int_0^t\sigma d\tau_2\ ,\] and similarly we can define $\tilde{\sigma}_{\tau_2}$. Moreover, integrating both run times, $\tau_1$, $\tau_2$ we have \[\tilde{\tilde{\sigma}}(\mathbf{x}_1,\mathbf{x}_2,t,\theta_1,\theta_2)=\int_0^t\int_0^t\sigma d\tau_1 d\tau_2\ .\] 

After stopping with rate given by $\beta_1$, robot $1$ starts a new run at $\tau_1=0$ which is described by
\begin{align}
\tilde{\sigma}_{\tau_1}(&\mathbf{x}_1,\mathbf{x}_2,t,\theta_1,\theta_2,{\tau_1=0})\nonumber \\  &=\int_Sk(\theta_1^*;\theta_1)\int_0^t\beta_1\tilde{\sigma}_{\tau_1}(\mathbf{x}_1,\mathbf{x}_2,t,\theta_1^*,\theta_2,\tau_1)d\tau_1d\theta_1^*\nonumber\\ & = T_1\int_0^t\beta_1\tilde{\sigma}_{\tau_1}(\mathbf{x}_1,\mathbf{x}_2,t,\theta_1,\theta_2,\tau_1)d\tau_1\ .\label{eq: initial condition of run}
\end{align}
 We can similarly define $\tilde{\sigma}_{\tau_2}(\mathbf{x}_1,\mathbf{x}_2,t,\theta_1,\theta_2,\tau_2=0)$.
 
 Integrating (\ref{eq: initial model}) with respect to $\tau_1$ and $\tau_2$ and using (\ref{eq: initial condition of run}) we obtain
 \begin{equation}
\begin{aligned}
{{\partial_{t}\tilde{\tilde{\sigma}}}}+{{c\theta_{1}\cdot\nabla_{\mathbf{x}_{1}}\tilde{\tilde{\sigma}}}}+{{c\theta_{2}\cdot\nabla_{\mathbf{x}_{2}}\tilde{\tilde{\sigma}}}}=&-{{(\mathds{1}- T_1)\int_0^t\int_0^t\beta_{1}\sigma d\tau_1d\tau_2}} \\&-{{(\mathds{1}- T_2)\int_0^t\int_0^t\beta_2\sigma d\tau_1d\tau_2}}\ .\label{eq: system}
\end{aligned}
\end{equation}

\subsection{PDE model for interacting robots with L\'{e}vy strategies}\label{sec: PDE model}

In  applications, the  mean run time $\bar{\tau}$ is often small compared with the macroscopic time scale $\mathcal{T}$  \textcolor{black}{in which the movement of the swarm of robots is observable}, and we aim to study (\ref{eq: system}) for \textcolor{black}{$\varepsilon=\bar{\tau}/\mathcal{T}\ll 1$} \cite{alt1980biased}. We introduce normalized variables $t_n,\ \mathbf{x}_n,\ \tau_n,$ and $c_n$ and a diffusion limit of (\ref{eq: system}) is obtained under the scaling $(\mathbf{x},t,\tau)\mapsto(\mathbf{x}_ns/\varepsilon,\thinspace t_n/\varepsilon,\thinspace  \tau_n/\varepsilon^\mu)$, with $c_n=\varepsilon^{-\gamma}c_0$
for $\mu,\ \gamma>0$. We further assume that the diameter of each robot is small,
$\varrho=\varepsilon^\xi$, \textcolor{black}{with $\xi>0$}, while the number of robots $N$ is large so that $(N-1)\varrho=\varepsilon^{\xi-\vartheta}$, with $\xi-\vartheta<0$.

\textcolor{black}{These assumptions are frequently satisfied in area coverage or search problems, where the area occupied by the robots is relatively small, but nonzero, compared to the area of the total arena. For example, in the experiments simulated in Section 6, $\varepsilon$ is of the order $0.005$. By applying the resulting equations to initial conditions localized around each robot, the resulting equations allow to approximate quantities of interest, like coverage, hitting times or optimal placement, also for finite robot diameter $\varrho$ \cite{zhang2018performance}.}

In the above parabolic scaling, this section obtains a fractional diffusion equation from  (\ref{eq: system}) for the density of robots moving according to the model in Section \ref{sec: movement strategy}. 

First, from the two-particle density equation (\ref{eq: system}) we aim to derive an equation for the one-particle density function
\begin{equation}
    p(\mathbf{x}_1,t,\theta_1)=\frac{1}{|S|}\int_0^t\int_0^t\int_{\Omega_2}\int_{S}\sigma d\theta_2d\mathbf{x}_2d\tau_1d\tau_2\ .\label{eq: density of 1}
\end{equation}
Here we follow some of the main steps in \cite{estrada2018swarming} where we derived a macroscopic PDE for  a system of interacting particles with long-range diffusion and alignment. 

We integrate (\ref{eq: system}) with respect to the accessible phase space $(\mathbf{x}_2, \theta_2)\in\Omega_2\times S$, where
$
\Omega_2=\Omega_2(\mathbf{x}_1)=\{\mathbf{x}_2\in\mathds{R}^2:\ |\mathbf{x}_1-\mathbf{x}_2|>\varrho  \} = \mathbb{R}^2\setminus B_\varrho(\mathbf{x}_1)
$ and $B_\varrho(\mathbf{x}_1)$ is a ball of radius $\varrho$, centered around $\mathbf{x}_1$. Introducing the scaling at the beginning of this section we obtain
\begin{equation}
\begin{aligned}
    \varepsilon\partial_tp&+\varepsilon^{1-\gamma} c_0\theta_1\cdot\nabla p=\varepsilon^{-\gamma}\frac{c_0}{|S|}\int_{\partial B_\varrho}\int_S\nu\cdot(\theta_1-\theta_2)\tilde{\tilde{\sigma}}d\theta_2d\mathbf{x}_2\\  &-(\mathds{1}-T_1)\int_0^t\mathcal{B}^\varepsilon(\mathbf{x}_1,t-s)p(\mathbf{x}_1-c\theta_1(t-s),s,\theta_1)ds\ .\label{eq: one particle density}
\end{aligned}
\end{equation}
The first term in the right hand side of the above expression describes the collision between the robots, while the second term describes the long-range movement, encoded in the operator $\mathcal{B}$. See Appendix \ref{sec: derivation} for the details of the derivation.\\

Up to lower order terms, we expand $p(\mathbf{x}_1,t,\theta_1)$ as follows \begin{equation}
p(\mathbf{x}_1,t,\theta_1)=|S|^{-1}(u(\mathbf{x}_1,t)+\varepsilon^\gamma 2\theta_1\cdot w(\mathbf{x}_1,t)+o(\varepsilon^\gamma)),\label{eq: expansion of p1}
\end{equation}
where 
$$u(\mathbf{x}_1,t)=\int_Spd\theta_1  \ \ \textnormal{and}\ \ w(\mathbf{x}_1,t)=\int_S\theta_1 pd\theta_1\ .
$$
The quantity $u(\mathbf{x}_1,t)$ represents a macroscopic density since it does not depend on individual characteristics of the robots such as the direction of motion $\theta$ or the run time $\tau$. $w(\mathbf{x}_1,t)$ describes the mean direction of a density of robots.

Substituting (\ref{eq: expansion of p1}) into (\ref{eq: one particle density}) and integrating with respect to $\theta_1$, we obtain the conservation law for the macroscopic density:
 \begin{equation}
     \partial_tu(\mathbf{x}_1,t)+ 2c_0\nabla\cdot w(\mathbf{x}_1,t)=0\ .\label{eq: conservation}
 \end{equation}
 Note that the first term in the right hand side vanishes due to the symmetry in $\theta_1$ and $\theta_2$ and in the second one we use (\ref{eq: conservation T}).

The final step is to compute the mean direction of motion $w$ and substitute it into the conservation equation. 

Starting with the collision term in (\ref{eq: one particle density}) we can write
\[
\tilde{\tilde{\sigma}}(\mathbf{x}_1,\mathbf{x}_2,t,\theta_1,\theta_2)=\tilde{\tilde{\sigma}}(\mathbf{x}_1,\mathbf{x}_1-\nu\varrho,t,\theta_1,\theta_2)
\]
since the normal vector $\nu$ at the time of collision is given by
$
\nu=(\mathbf{x}_1-\mathbf{x}_2)/\varrho$ hence, $\mathbf{x}_2=\mathbf{x}_1-\nu\varrho$. The key step to re-write this term is to use the molecular chaos assumption, which is plausible at low density of robots \cite{cercignani2013mathematical,franz2016hard}, where collisions between more that $2$ robots is neglected. It states that the velocity of the robots is approximately independent of each other, so that $\tilde{\tilde{\sigma}}$ approximately factors into one-particle densities $
 \tilde{\tilde{\sigma}}(\mathbf{x}_1,\mathbf{x}_1\pm\varepsilon^{\xi}\nu,t,\theta_1,\theta_2)=p(\mathbf{x}_1,t,\theta_1)p(\mathbf{x}_1,t,\theta_2)+\mathcal{O}(\varepsilon^\xi)$.

For the second term in the right hand side of (\ref{eq: one particle density}) we use a quasi-static approximation such that $\hat{\mathcal{B}}^\varepsilon(\mathbf{x}_1,\varepsilon\lambda+\varepsilon^{1-\gamma}c_0\theta_1\cdot\nabla)\simeq\hat{\mathcal{B}}^\varepsilon(\mathbf{x}_1,\varepsilon^{1-\gamma}c_0\theta_1\cdot\nabla)$, since $\gamma>0$ and then we obtain a term involving the fractional Laplacian as described in Appendix \ref{app: fractiona laplacian}.

Finally, rewriting the right hand side of (\ref{eq: one particle density}) as described above we multiply by $\theta_1$ and integrate in $S$ to obtain the following result.
\begin{theorem}\label{thm: parabolic limit}
 As $\varepsilon\rightarrow 0$, the macroscopic density $u(\mathbf{x},t)$ satisfies the following fractional diffusion equation:
\begin{equation}\label{eq:final}
\partial_tu=c_0\nabla \cdot\left( \frac{1}{F(u)} \ C_\alpha\nabla^{\alpha-1}u \right)
 \end{equation}
 \begin{align}
F(u)&=\frac{\alpha-1}{\varsigma_0|S|}(1-\nu_1)+\frac{32c_0^3}{3|S|^2}u,\label{F}\\
 C_\alpha &=-\frac{\varsigma_0^{\alpha-2}c_0^{\alpha-1}(\alpha-1)^2\pi}{\sin(\pi\alpha)\Gamma(\alpha)}\frac{(|S|-4\nu_1)}{|S|^2}\ .\label{C}
 \end{align}
The term $F(u)$ comes from the interactions and $C_\alpha$ is the diffusion coefficient. 
 \end{theorem}
\textcolor{black}{\begin{proof}
 See Appendix \ref{app: fractiona laplacian} for the derivation of the fractional Laplacian and \cite{estrada2018swarming} for further details on the proof of the Theorem.
\end{proof}}
 
 To measure the coverage of the domain in the continuum sense, and be able to compare it with the discrete robotic simulations we define the following time averaged coverage function, used later in Section \ref{sec: comparison}.
\begin{equation}\label{eq:coverage}
\textrm{Cov}(t) = \frac{1}{t} \int_0^t \int_{\Omega} \min(u(\mathbf{x},s),\bar{\rho}) d\mathbf{x} ds\ \ \textnormal{where}\ \ \bar{\rho}=\frac{1}{|\Omega|}\ .
\end{equation}
\textcolor{black}{The idea for this definition of coverage is as follows: Once the density $u$ reaches a threshold $\bar{\rho}$ it is considered to be covered. To avoid over-counting already covered areas, the min function is used. Since $\int_{\Omega} u d\mathbf{x}$ is constant for all time $t\geq 0$, a natural choice for the threshold is the uniform distribution of the macroscopic density in $\Omega$.}


\textcolor{black}{The second quantity of interest which we compare with the \textcolor{black}{individual robot} simulations is the expected hitting time (see \cite{schroeder2017efficient} for analogous definition of hitting time). For a given target, this is defined as the time taken before a robot reaches this target. For the target $\mathsf{T} \in \Omega$ occupying a volume $\mathrm{vol}(\mathsf{T})>0$ we seek the hitting time $t_0 \geq 0$ at which the density of the solution $u$ reaches certain threshold $\delta$, i.e.,
\[
\delta=\int_{\mathsf{T}} u(\mathbf{x},t_0)d\mathbf{x}\ .
\]
In Section \ref{sec: comparison} we use an analytic expression for the solution $u$ in $\mathds{R}^n$ obtained in \cite{m3as} as well as a numerical solution.}


\section{Robot simulations}\label{sec: simulation of robots}

\begin{figure}[!htp]
    \centering
    \subfloat{\includegraphics[width=0.45\textwidth]{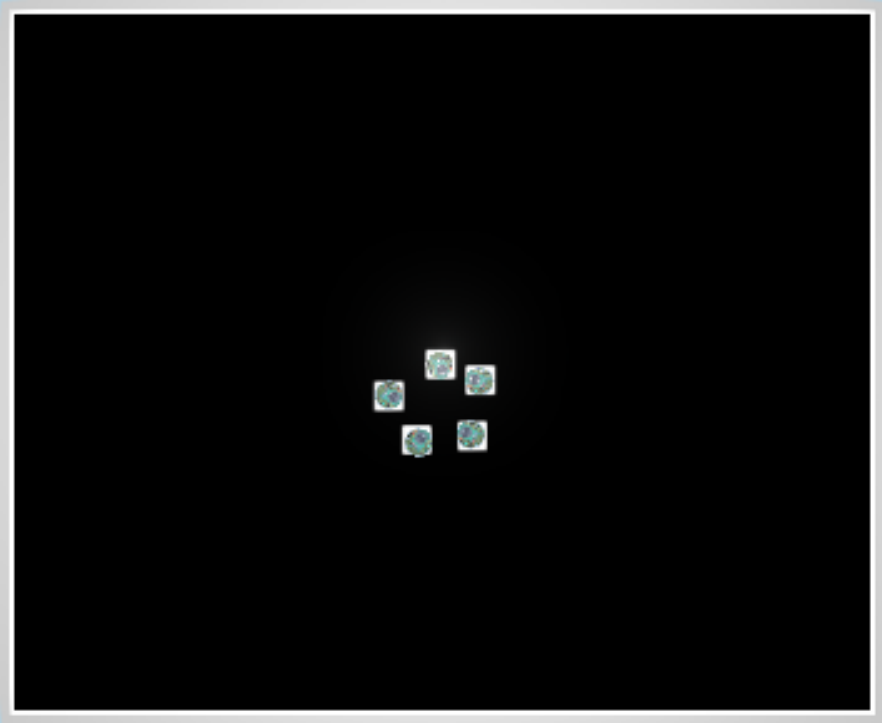}}
    \subfloat{\includegraphics[width=0.45\textwidth]{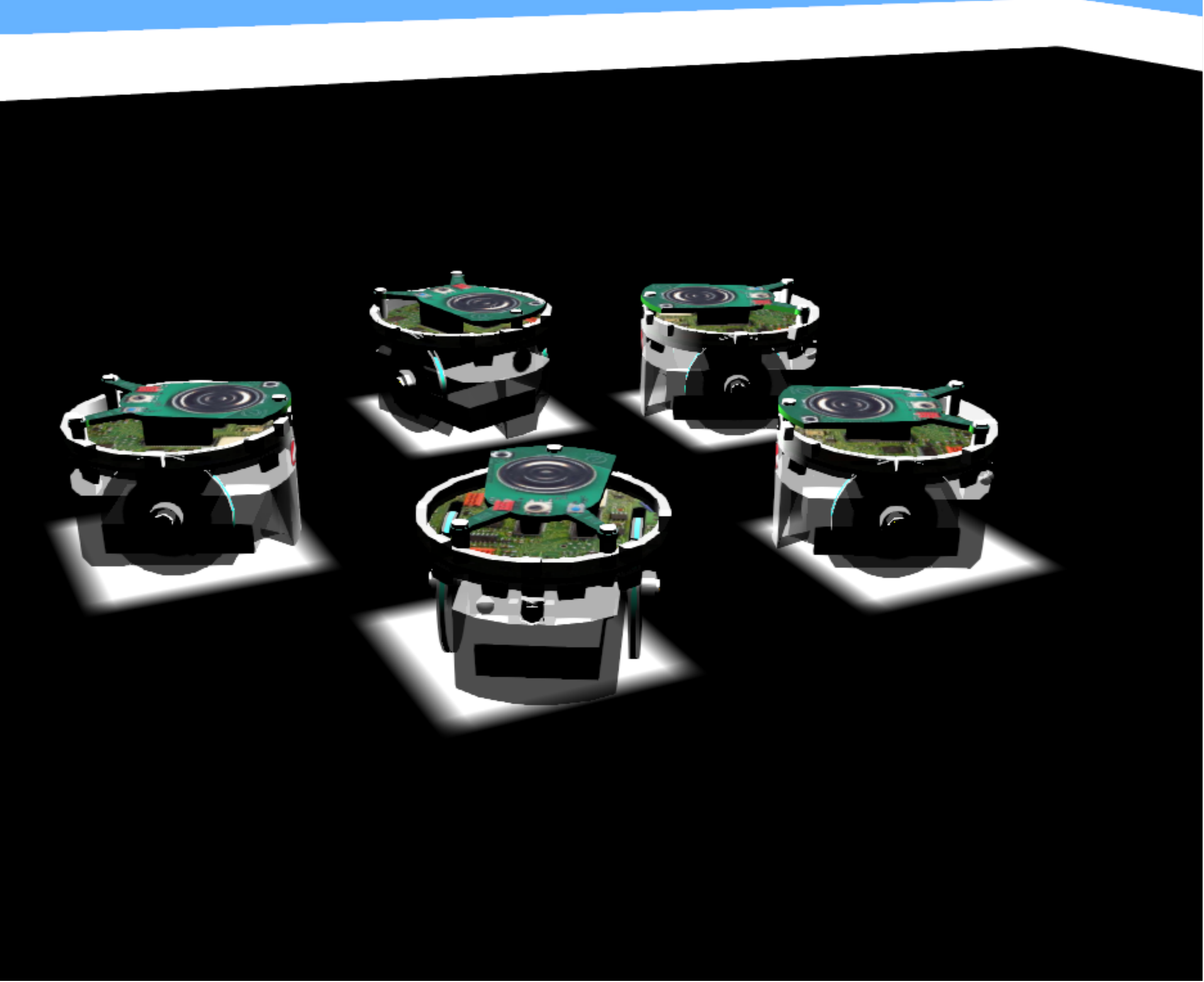}}
     \caption{These two images show the placement of the \textit{e-puck} robots within the arena for all \textcolor{black}{simulations} with 5 robots in the Webots \cite{webots} simulator. On the left, a birds-eye view of the initial positions around the centre of the arena can be seen - the robots are placed in a ring formation. On the right, a zoomed in side view of the initial orientations in the same arena - the robots are facing away from each other.}\label{fig:epuckRobotsInitial}
\end{figure}

A L\'{e}vy search strategy was implemented for the \textit{e-puck} \cite{epuck} robot using the Webots \cite{webots} 2019a simulator in a arena of dimensions 220cm by 180cm. Each \textit{e-puck} is a differential wheeled robot with a circular body with a diameter of 7.5cm and 2 wheels of a diameter of 4.1cm. The robots are initially placed at centre of the simulated arena. Specifically, they are placed in a ring and oriented to face away from one another. For \textcolor{black}{simulations} of robot population sizes 5,10,15 and 20 the diameter of these initial positions are 25cm, 30cm, 40cm and 55cm. An example for the initial setup with 5 robots is shown in Figure \ref{fig:epuckRobotsInitial}.

\begin{figure}[!htp]
    \centering
\includegraphics[width = 0.45\textwidth]{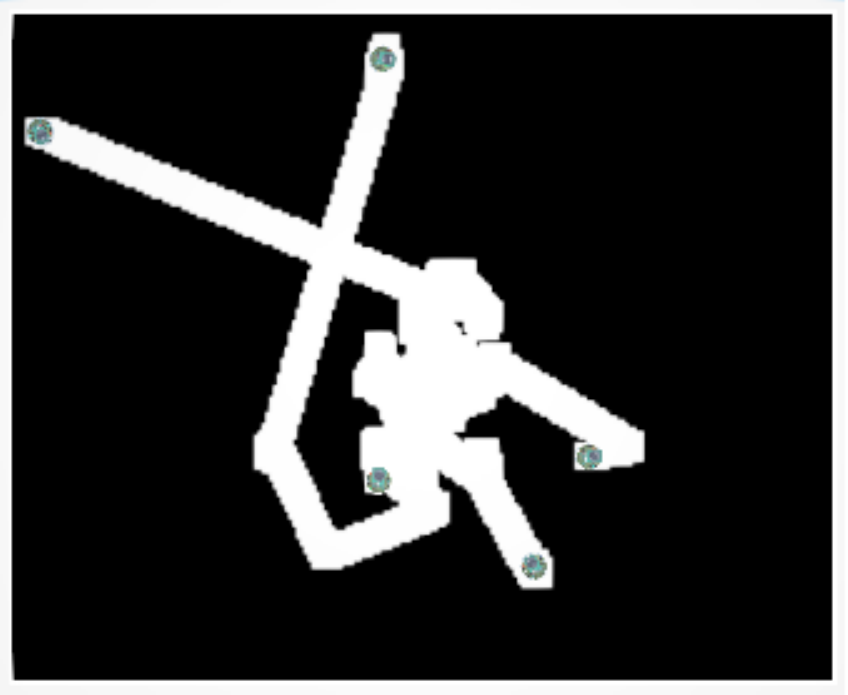}
     \caption{This image shows a birds-eye view of the coverage trails left by 5 \textit{e-puck} robots within the arena in the Webots \cite{webots} simulator after two minutes of execution. The \textit{e-puck} robot's (small green disk) coverage can be visualised by overlaying an image onto the base of the arena. It can be seen that the robots have left a cluster of small trails around their initial centre positions, seen in Figure \ref{fig:epuckRobotsInitial}, before performing slightly larger movements which have spread the robots out within the arena.}\label{fig:epuckRobots}
\end{figure}

To track the coverage of the simulated arena, we discretized the environment by overlaying a virtual grid of $1\textnormal{cm}\times 1\textnormal{cm}$ cells onto it. Each cell captures the simulation time at which it was reached by any robot (hit time) and therefore the coverage at any one time can be computed by simply counting the number of non-empty cells. Figure \ref{fig:epuckRobots} shows the simulator's display tool used to visualize visited cells (in white) and empty cells (in black).

The coverage grid is maintained by a supervisor agent within the simulation, which is able to access the position of all robots at each step of the simulation.
The supervisor records these positions into the coverage grid. In addition it regularly computes and saves the current coverage value. 

The supervisor computes the coverage as follows \cite{fossum2014repellent}:
\begin{equation}
    \textnormal{Coverage} = \frac{\textnormal{number of explored cells}}{\textnormal{total number of cells}} \ .
    \label{eq:coverageCalculation}
\end{equation}


The controller used by each \textit{e-puck} robot is shown in Algorithm \ref{EpuckAlgorithm}\footnote{Code available at 
}.

\begin{algorithm}
\SetAlgoLined
 \While{True}{
  $(x_{\textnormal{new}}, y_{\textnormal{new}})$ = l\'{e}vyDistribution()\\
  distanceToTravel = norm($(x_{\textnormal{new}}, y_{\textnormal{new}})$)\\ 
  angle = atan2(($x_{\textnormal{new}}, y_{\textnormal{new}}$))\\
  rotateRobot(angle)\\
  \While{distanceTraveled $<$ distanceToTravel}{
  	\eIf{no obstacles are detected}{
     moveForward()
     }{
     break
    }
  }
 }
 \caption{\textit{e-puck} L\'{e}vy Search controller. This algorithm outlines the behaviour of the \textit{e-puck} robots. The robot's new position is calculated within the \code{l\'{e}vyDistribution()} function and calculations required for this are shown in (\ref{eq:epuckMovement}).}
 \label{EpuckAlgorithm}
\end{algorithm}

The robots alternate on-the-spot turns (with rotational velocity fixed at 0.858 rad/sec) with straight movements (at 6.44 cm/sec) in the direction of target points that are computed using the L\'{e}vy distribution relative to their current position. \textcolor{black}{Robots do not avoid obstacles, pro-actively. Rather, they} stop and compute a new target point whenever their on-board Infrared (IR) range sensors detect an obstacle (either another robot or the borders of the arena).




The movement of each \textit{e-puck} is characterized by long distance runs, where the run distance $r$ is generated from a L\'{e}vy process \textcolor{black}{\cite{chambers1976method,harris}}
\textcolor{black}{\begin{align}
    r=\frac{\sin(\alpha\tilde{X}_1)}{(\cos(\tilde{X}_1))^{\nicefrac{1}{\alpha}}}\Bigl(\frac{\cos((1-\alpha)\tilde{X}_1)}{\tilde{X}_2} \Bigr)^{\frac{1-\alpha}{\alpha}}\ ,
    \label{eq:epuckMovement}
\end{align}}
and the new positions are computed, for $\theta=\pi X_3$, as
\begin{equation}
         x_{\textnormal{new}}-x_{\textnormal{current}}=r\cos(\theta),\ \textnormal{and} \ \ y_{\textnormal{new}}-y_{\textnormal{current}}=r\sin(\theta)\ .
\end{equation}



Here, $\tilde{X}_1=\nicefrac{1}{\pi}X_1$ is a uniform random variable in the interval $[-\nicefrac{\pi}{2},\ \nicefrac{\pi}{2}]$, $\tilde{X}_2=-\ln X_2$ has a unit exponential distribution and $X_1=X_2=X_3$ are uniformly distributed random variables in the interval $(0,1)$. Since we are calculating the new location ($x,y$) relative to the base of the robot, we can set 
    $x_{\textnormal{current}} = y_{\textnormal{current}} = 0$.


Each \textcolor{black}{simulation} is run for $T=20$ minutes and coverage values are recorded every second. The supervisor agent then resets the robots into their original positions, clears the coverage grid and saves experiment data to a file. These \textcolor{black}{simulations} are run in batches so that data for hundreds of different runs can be captured and compared. \textcolor{black}{Simulations} were run for different parameters;  $N= 5, 10, 15, 20$; and $\textcolor{black}{\alpha} = 1.1, 1.3, 1.5, 1.7, 1.9$. \textcolor{black}{Webots simulations include stochastic features such as sensor noise, motor noise and random alignment, unlike the macroscopic simulations in which this noise is averaged.}

\section{Comparison between macroscopic model and individual robot simulations}\label{sec: comparison}


\textcolor{black}{Numerical simulations of \eqref{eq:final}, considering $F(u)=1$, are based on a finite element approximation of the nonlocal partial differential equation, see for example \cite{Acosta1,Acosta2}.\\
We consider the weak formulation
\begin{equation}
    \int_0^T \langle \partial_t u, v \rangle - \langle c_0 {C}_{\alpha} \nabla^{\alpha-1} u,\nabla v \rangle dt + \langle u_0, v(0) \rangle = 0  ,
\end{equation}
for $u,v \in W(0,T) = \lbrace v \in L^2(0,T;H^{\alpha/2}(\Omega)): \partial_t v \in L^2(0,T;H^{-\alpha/2}(\Omega))\rbrace$ where $H^{\alpha/2}(\Omega)$ is a Sobolev space of smoothness $\alpha/2$. The bilinear form $ \langle \nabla^{\alpha-1} u,\nabla v \rangle$ is implemented using quadrature methods commonly used for singular integral operators \cite{Acosta1,Acosta2,VIpreprint}.\\
The fully discrete time stepping scheme reads as follows: Find $\lbrace u_h^1,u_h^2,\dots \rbrace \in W_h(0,T) \subset W(0,T)$  such that
\begin{align*}
    \left( M - \triangle t \ A \right) u_h^{n+1} = M\ u_h^n,
\end{align*}
where $u_h^0$ is given. $M,\ A$ are the mass and stiffness matrices related to the piecewise linear basis functions $\phi_i$ of $H_h \subset H^{\alpha/2}(\Omega)$ defined by $M_{ij}=(\phi_i,\phi_j)$, $A_{ij} = \langle \nabla^{\alpha-1} \phi_i,\nabla \phi_j \rangle$.\footnote{Code available at \iffalse http://www.macs.hw.ac.uk/\textasciitilde{}js325/robotic\textunderscore{}simulation/ \fi}.}\\

 We consider a robotics arena given by $\Omega = \left[-0.9,0.9 \right] \times \left[ -1.1, 1.1 \right]$ to match the Webots simulation. The initial condition for the continuum model \eqref{eq:final} is $u_0(x) = \max \left( 0, 1.2 \exp{\frac{-4 N \vert x\vert^2}{0.075}}-0.2 \right)$.  It approximates the initial condition for the individual \textit{e-puck} Webots simulations described in Section~\ref{sec: simulation of robots}. 

\textcolor{black}{From the physical characteristics of the robots we determine the values of the model parameters in Section 4.2. To do so, recall that the diameter of each robot is $\varrho=7.5\ \textnormal{cm}$, and it moves with a speed $c_n=3\ \textnormal{cm/s}$. As the scale $s$ is of order  $\textnormal{cm/s}$, from the dimensions of $\Omega$ we  obtain a value of $\varepsilon=0.005$ from $\mathbf{x}_n=\varepsilon\mathbf{x}/s$. We further note $\gamma = \frac{1}{2}$ obtain $c_0 = 3 \cdot 0.005^\gamma$ from the definition $c_n=\varepsilon^{-\gamma}c_0$.
These values of the parameters are in agreement with the assumptions in Section 4.2, and in particular $\varepsilon \ll 1$. Note that the more detailed properties of the robots, such as the fact that they are differential-wheeled, do not enter into the current mathematical model. The model only captures on-the-spot turns, independently of the physical characteristics of the wheels.}

\begin{figure}
    \centering
    \includegraphics[width = 0.5\textwidth]{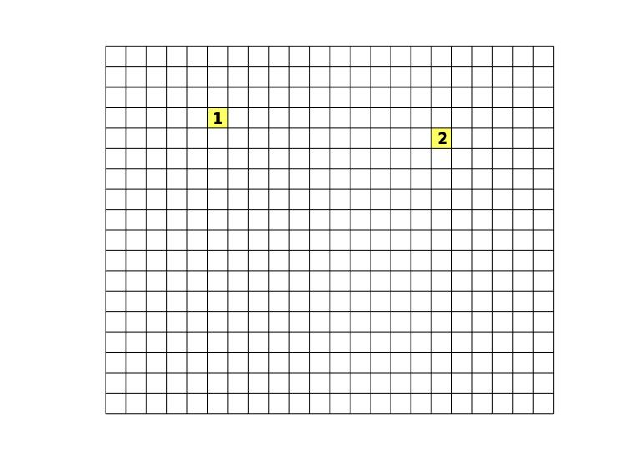}
    \caption{Domain for macroscopic simulation as well as Webots individual robot simulation. Tiles 1 and 2 are highlighted for convenience later.}
    \label{fig:domain}
\end{figure}

\begin{figure}
    \centering
    \subfloat[$N=5$]{\includegraphics[width = 0.49\textwidth]{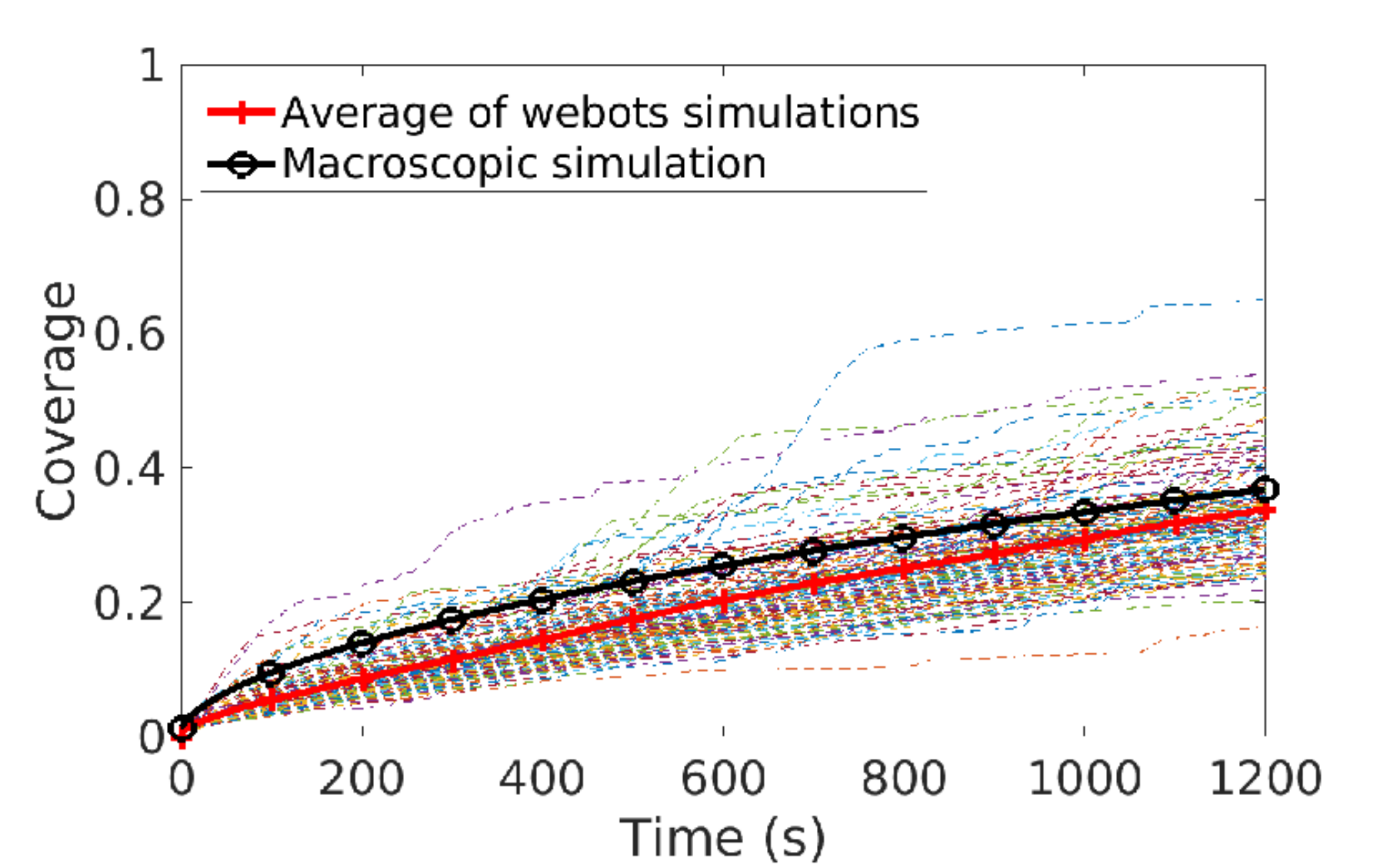}}
    \subfloat[$N=10$]{\includegraphics[width = 0.49\textwidth]{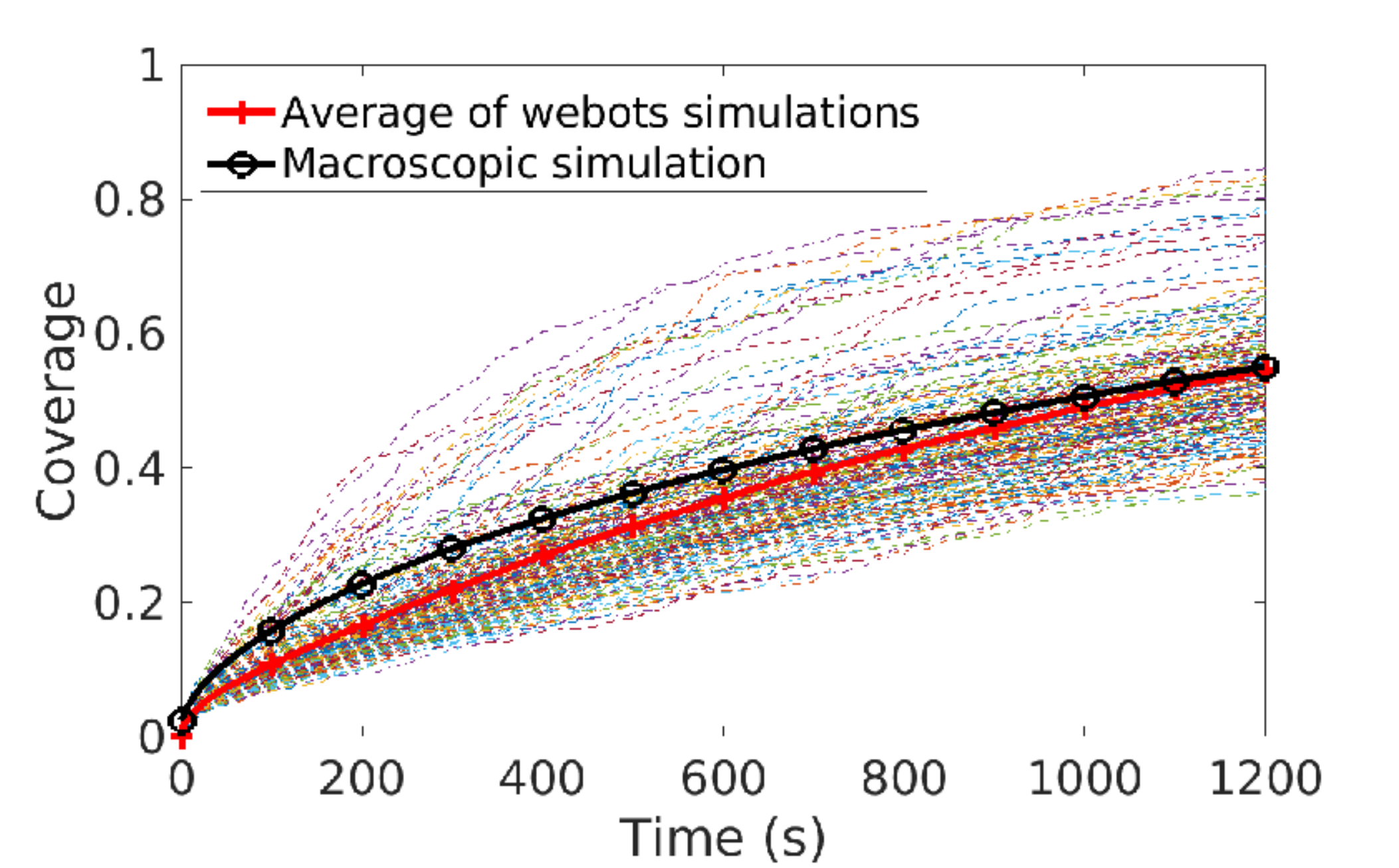}}\\
    \subfloat[$N=15$]{\includegraphics[width = 0.49\textwidth]{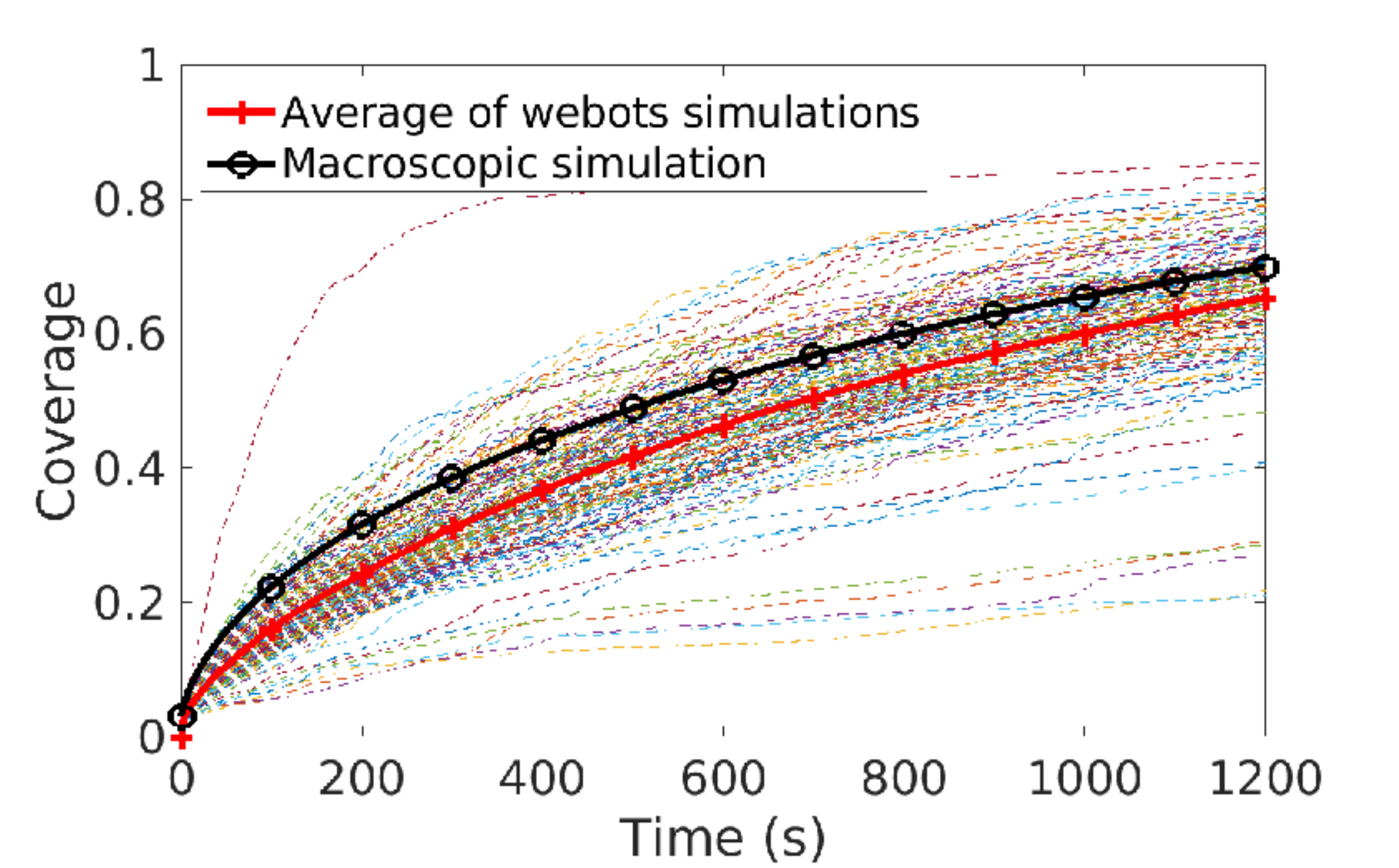}}
    \subfloat[$N=20$]{\includegraphics[width = 0.49\textwidth]{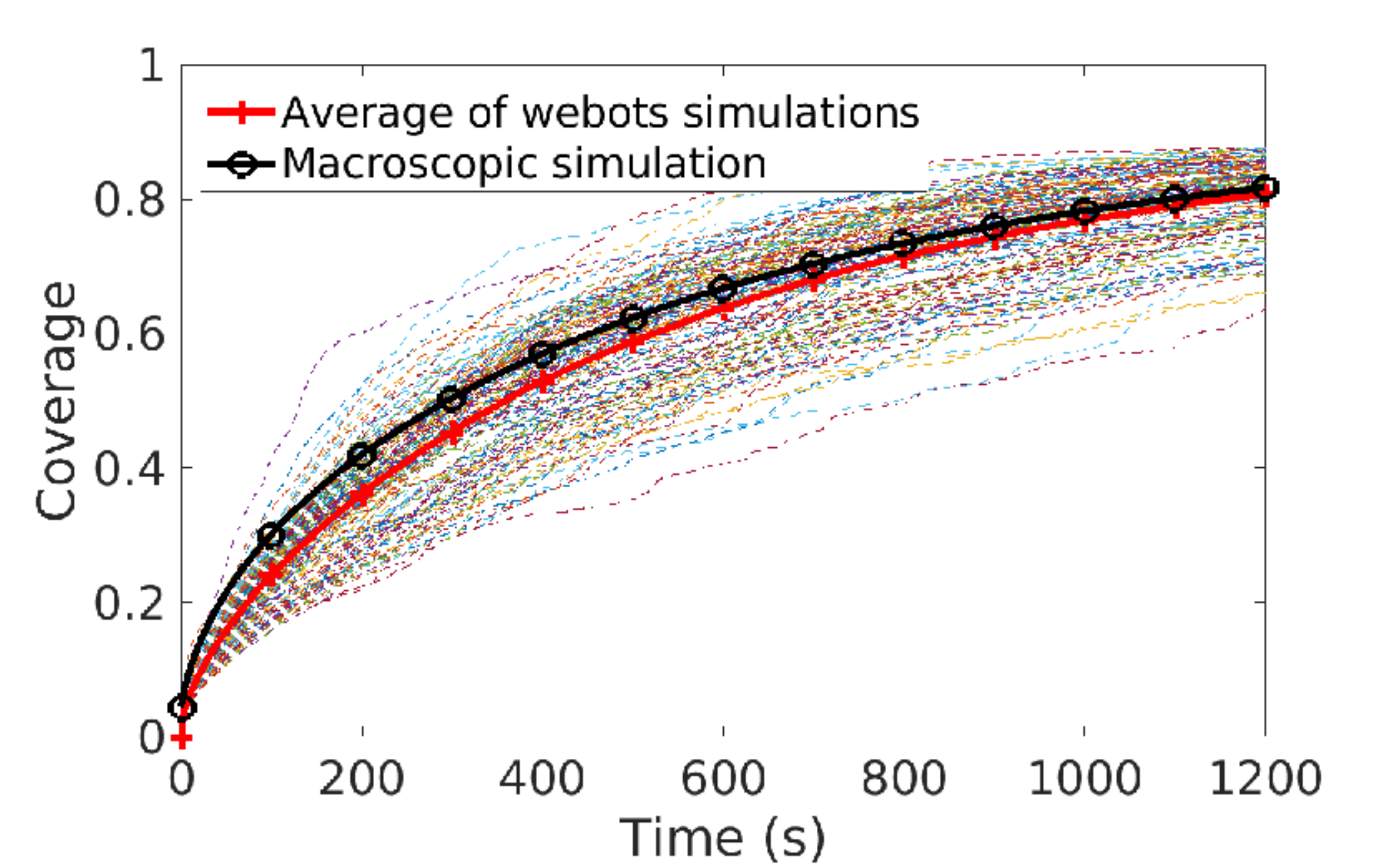}}
    \caption{Comparison of the Webots simulations with the macroscopic model for $\alpha = 1.3$ and varied number of robots $N=5,10,15,20$. \textcolor{black}{Dashed lines represent individual Webots runs.}}
    \label{fig:comp_a13_Nvaried}
\end{figure}
\begin{figure}
    \centering
    \subfloat[$\alpha = 1.1$]{\includegraphics[width = 0.49\textwidth]{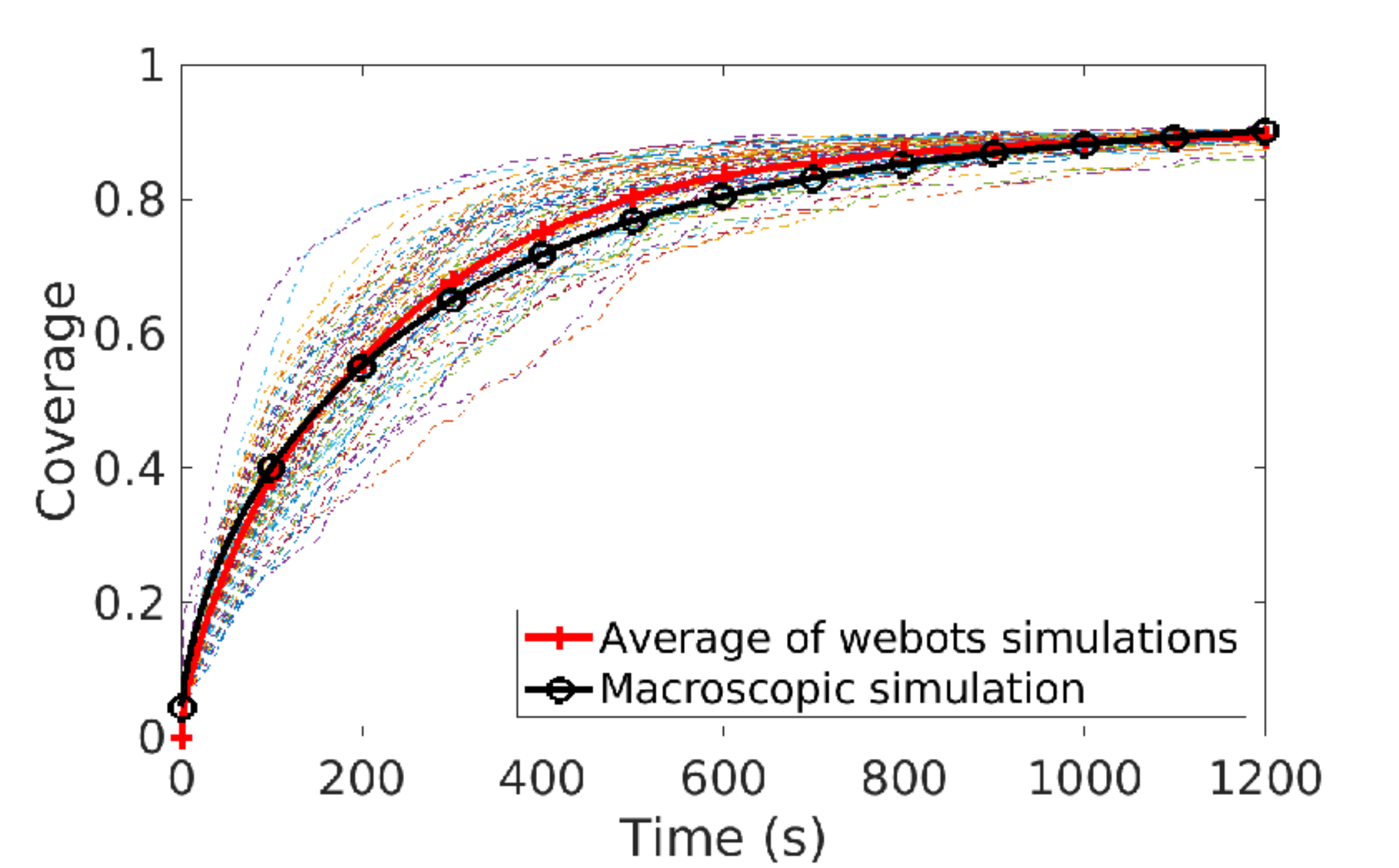}}
    \subfloat[$\alpha = 1.3$]{\includegraphics[width = 0.49\textwidth]{N20_A13_FF2_vF.pdf}}
    
    \subfloat[$\alpha = 1.5$]{\includegraphics[width = 0.49\textwidth]{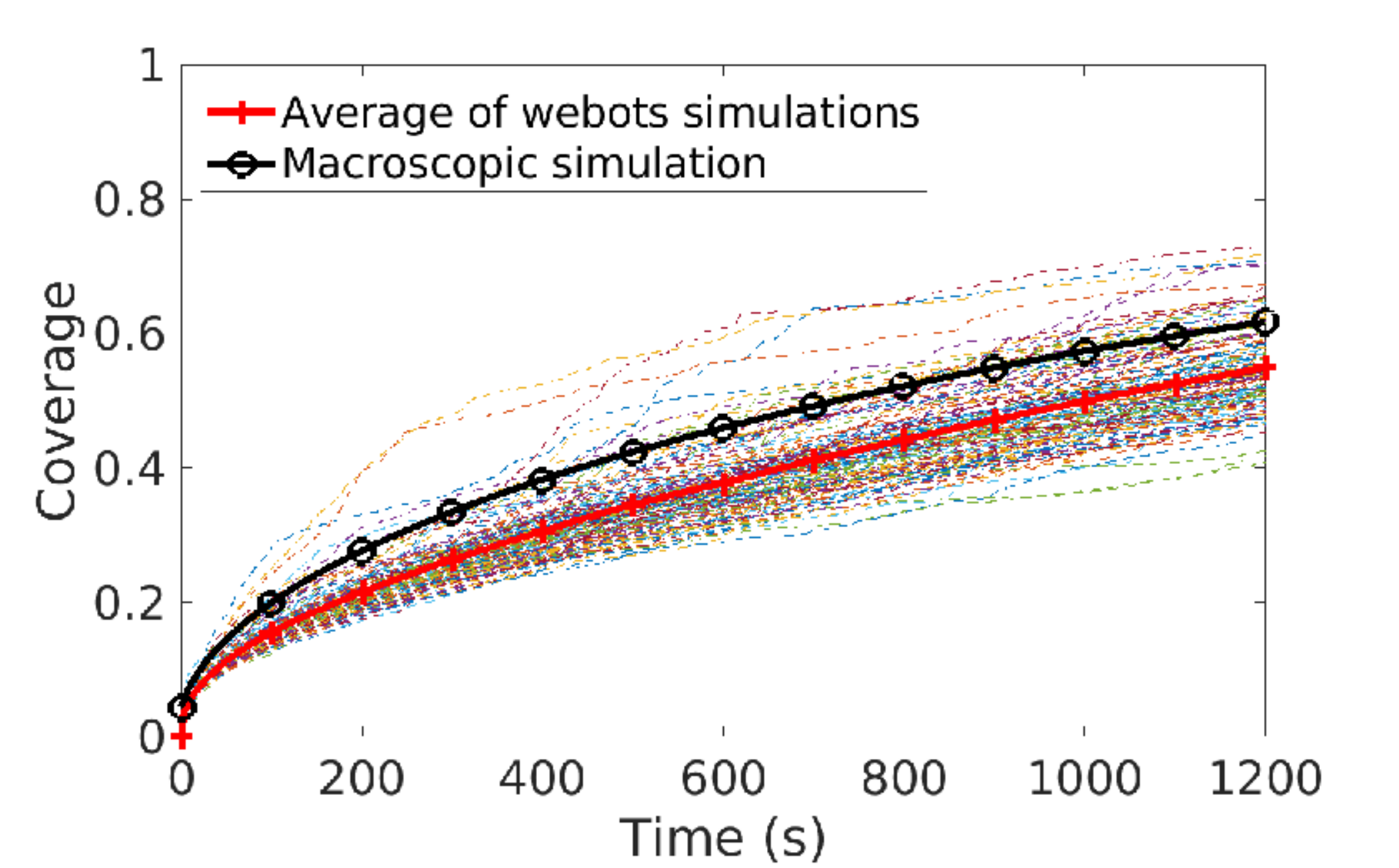}}
    \subfloat[$\alpha = 1.7$]{\includegraphics[width = 0.49\textwidth]{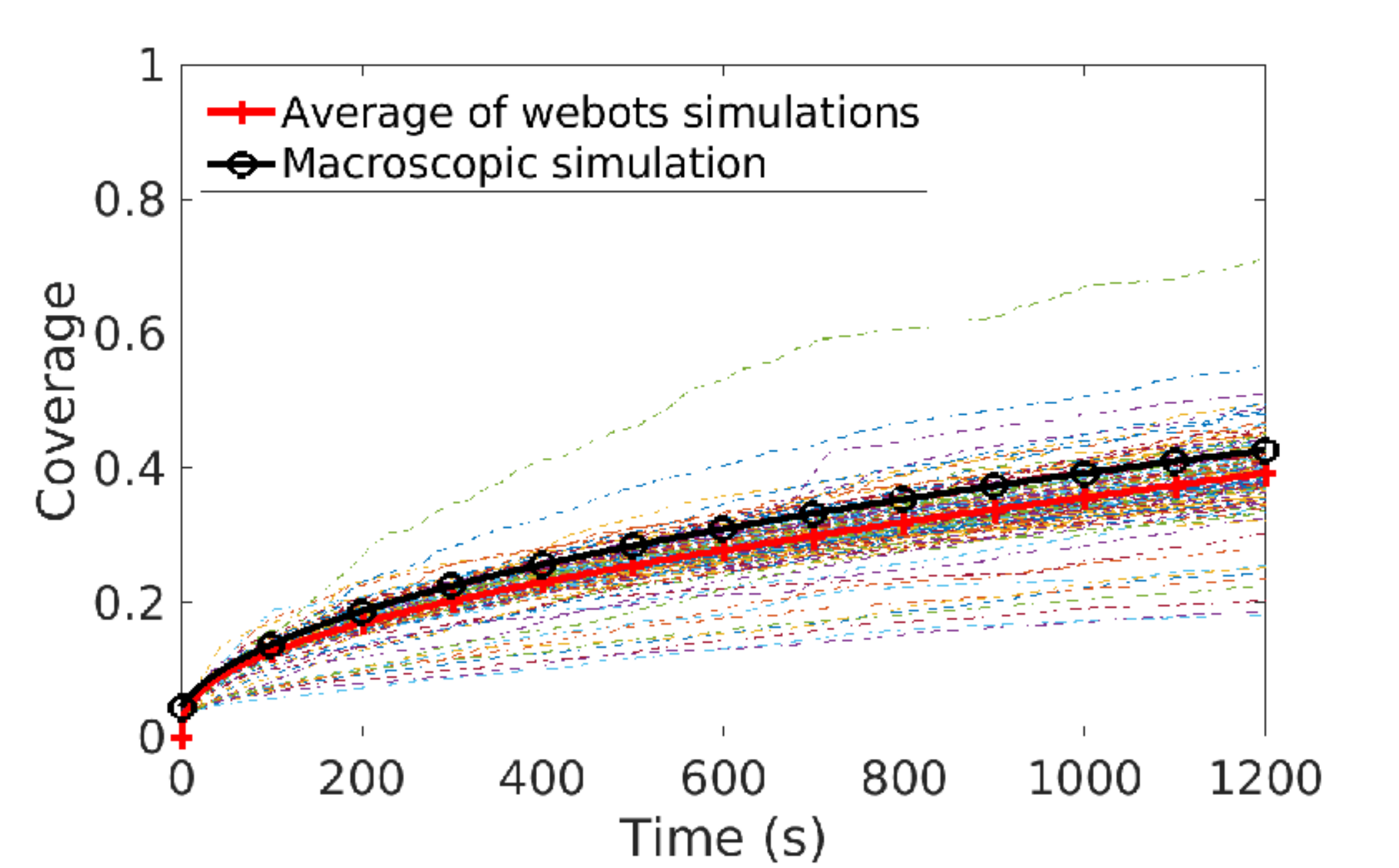}}
    
    \subfloat[$\alpha = 1.9$]{\includegraphics[width = 0.49\textwidth]{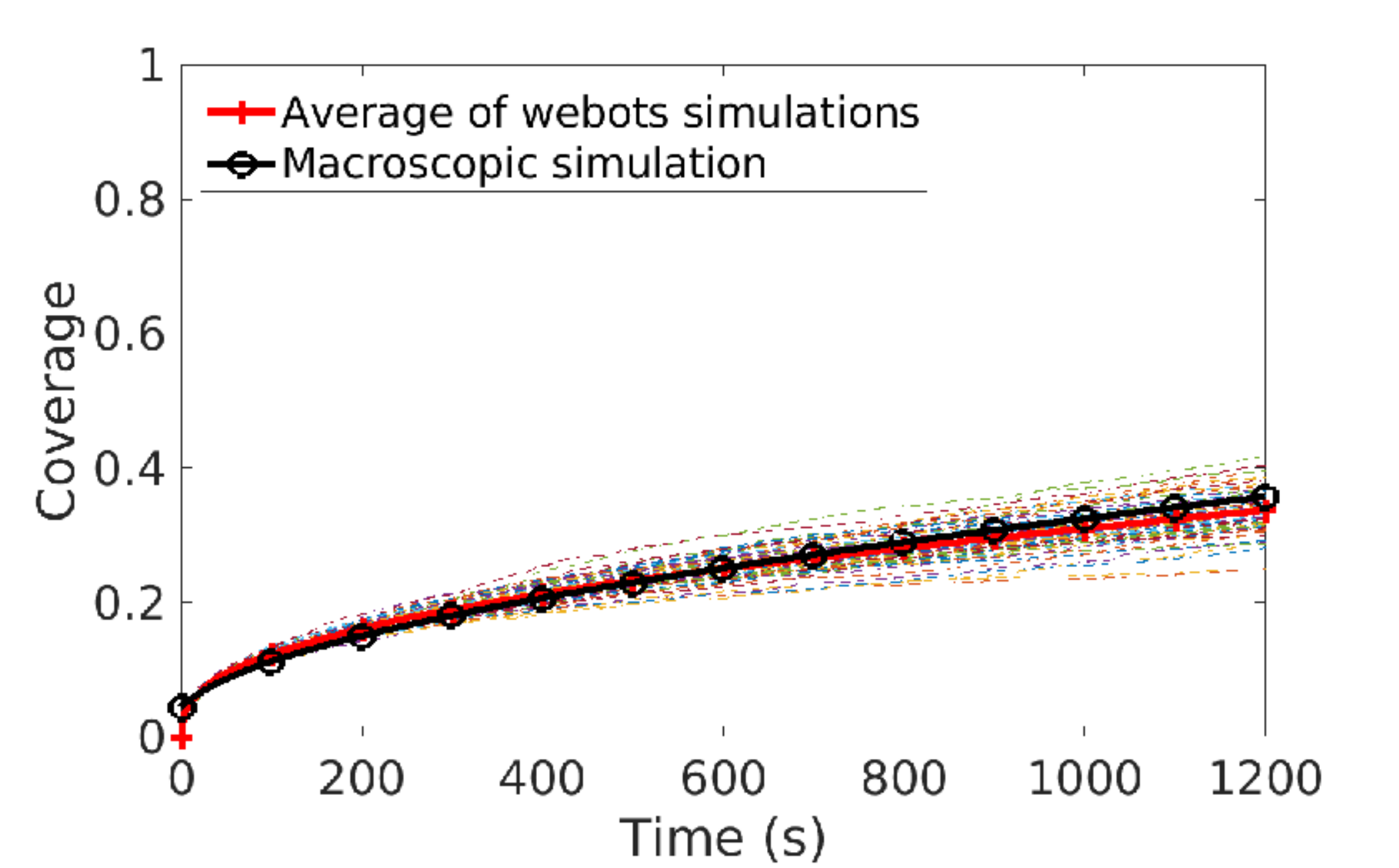}}
    \caption{Comparison of the Webots simulation with the macroscopic model for $N=20$ and various $\alpha = 1.1,\ 1.3,\ 1.5,\ 1.7,\ 1.9$. \textcolor{black}{Dashed lines represent individual Webots runs.}}
    \label{fig:comp20}
\end{figure}

Figure \ref{fig:comp_a13_Nvaried} compares PDE and Webots individual agent simulations for different numbers of robots, $N=5,\ 10,\ 15,\ 20$ for a  L\'{e}vy exponent $\alpha = 1.3$. \textcolor{black}{Dashed lines represent individual Webots runs.} The coverage increases with time and $N$, and the average of the Webots simulations agrees closely with the PDE solution. Figure~\ref{fig:comp20} similarly compares the coverage as a function of $\alpha$, with $N=20$, and finds close agreement between the PDE simulations and the average of Webots simulations. We note that coverage increases with decreasing $\alpha$, showing the advantage of long-range L\'{e}vy strategies. \textcolor{black}{Again dashed lines represent individual Webots runs.}\\
In all cases, as the number of robots $N$ increases we observe an increase in coverage efficiency as well as a decrease in the variation of the individual runs. In the transient regime far from full coverage, the variations are significantly larger for the longer-ranged L\'{e}vy strategies.  \\


\begin{figure}
    \centering
    \includegraphics[width = 0.6\textwidth]{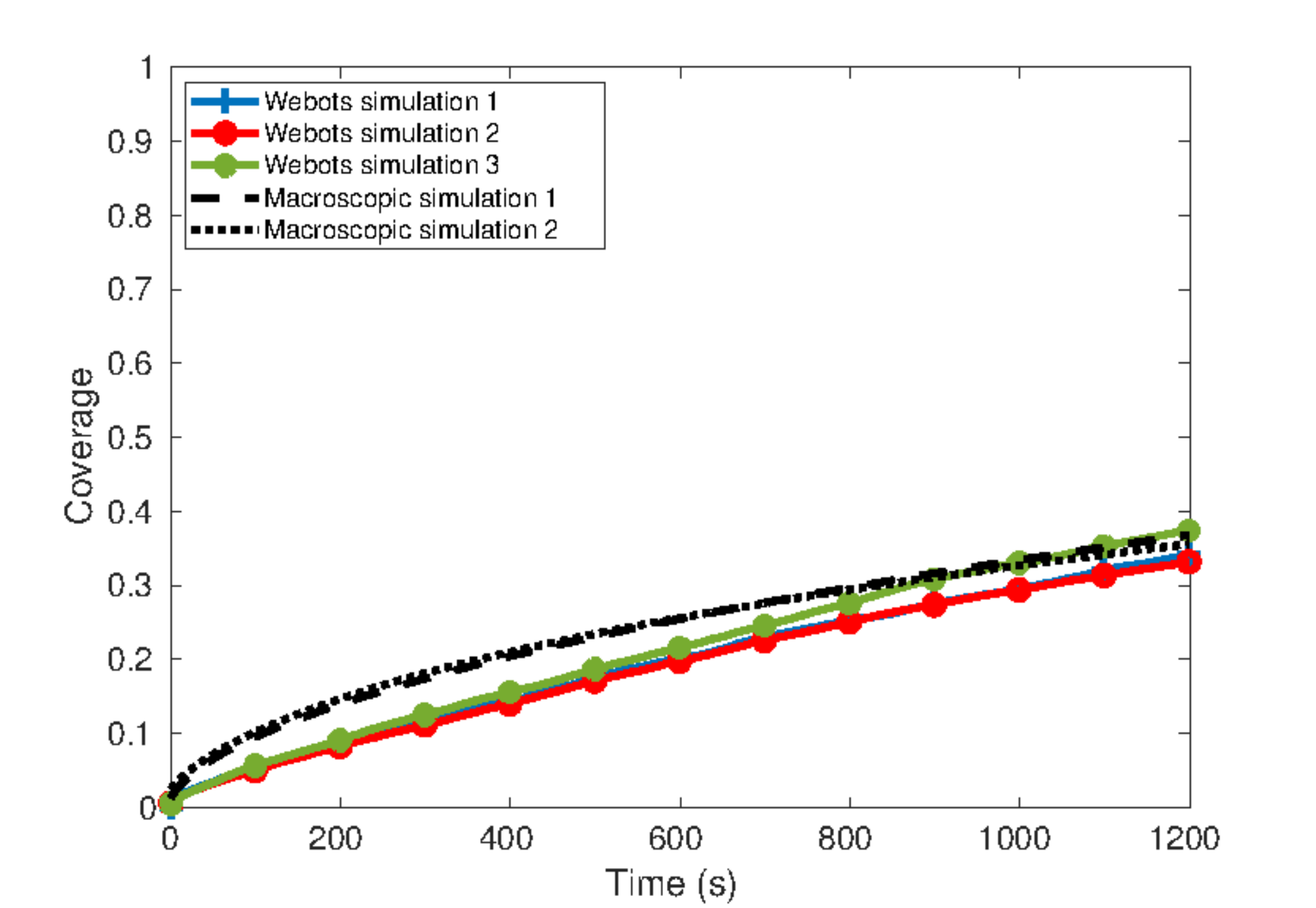}
    \caption{Comparison of Webots simulations and macroscopic simulations for different initial conditions for $N=5$ robots and $\alpha = 1.3$. The Webots simulations 1--3 depict an average of 100 runs. For Webots simulation 1 the robots are initially placed in a ring of diameter $25$cm oriented away from the center. Webots simulation 2 begins with robots placed in a ring of diameter $25$cm oriented along the $x$-axis. Webots simulation 3 begins with robots placed locations $(0,0),\ (0.9,0.7),\ (0.9,-0.7),\ (-0.9,0.7),\ (-0.9,-0.7)$ in the arena and oriented along the $x$-axis. Macroscopic simulation 1 corresponds to the initial condition $u_0(x) = \max \left( 0, 1.2 \exp{\frac{-4 N \vert x\vert^2}{0.075}}-0.2 \right)$ while Macroscopic simulation 2 corresponds to the initial condition $u_0(x) = \sum_{i=1}^5 \max \left( 0, 1.2 \exp{\frac{-20 \vert x-x_0^{(i)}\vert^2}{0.075}}-0.2 \right)$ with $x_0^{(1)} = (0,0),\ x_0^{(2)} = (0.9,0.7),\ x_0^{(3)} = (0.9,-0.7),\ x_0^{(4)} = (-0.9,0.7),\ x_0^{(5)} = (-0.9,-0.7)$.}
    \label{fig:Diff_ICs}
\end{figure}

\textcolor{black}{We present a comparison between Webots simulations and macroscopic simulations for different initial conditions for $N=5$ robots and $\alpha = 1.3$ in Figure~\ref{fig:Diff_ICs}. The Webots simulations 1--3 depict an average of 100 runs. Webots simulation 1 begins with robots placed in a ring of diameter $25$cm oriented away from the center as described in Section~\ref{sec: simulation of robots}. Webots simulation 2 begins with robots placed in a ring of diameter $25$cm oriented along the $x$-axis. Webots simulation 3 begins with robots placed locations $(0,0),\ (0.9,0.7),\ (0.9,-0.7),\ (-0.9,0.7),\ (-0.9,-0.7)$ in the arena and oriented along the $x$-axis. Macroscopic simulation 1 corresponds to the initial condition $u_0(x) = \max \left( 0, 1.2 \exp{\frac{-4 N \vert x\vert^2}{0.075}}-0.2 \right)$ while Macroscopic simulation 2 corresponds to the initial condition $u_0(x) = \sum_{i=1}^5 \max \left( 0, 1.2 \exp{\frac{-20 \vert x-x_0^{(i)}\vert^2}{0.075}}-0.2 \right)$ with $x_0^{(1)} = (0,0),\ x_0^{(2)} = (0.9,0.7),\ x_0^{(3)} = (0.9,-0.7),\ x_0^{(4)} = (-0.9,0.7),\ x_0^{(5)} = (-0.9,-0.7)$. Figure~\ref{fig:Diff_ICs} shows the coverage as a function of time. We observe that for macroscopic times the different initial conditions lead to a similar coverage, given the number of robots. This is expected for a random walk model described by a macroscopic diffusion equation.}

\begin{figure}
    \centering
\includegraphics[width = 0.75\textwidth]{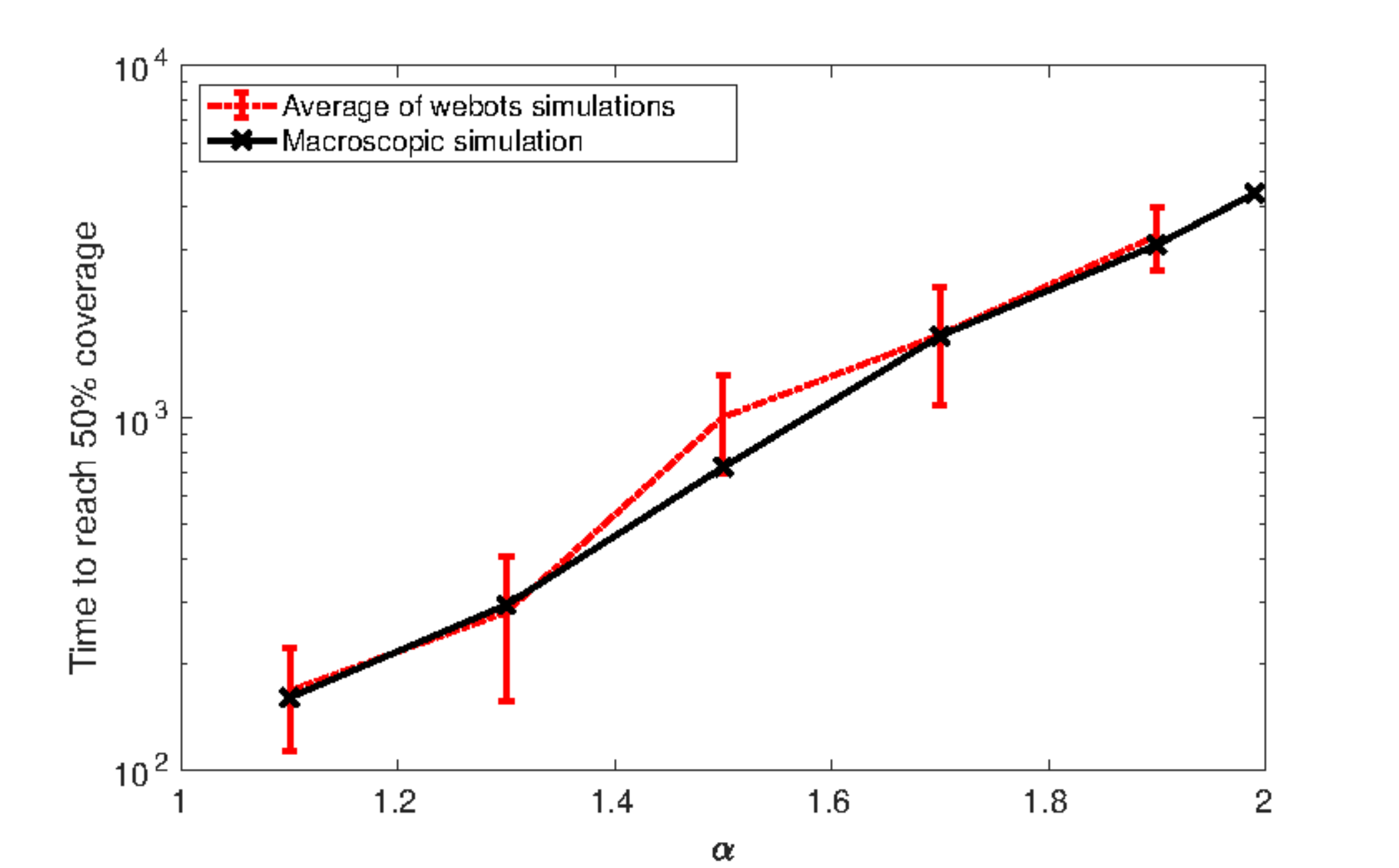}
    \caption{Comparison of average time to reach $50\%$ coverage for various values of $\alpha$ and $N=20$ robots. {The error bars show one standard deviation of the Webots simulations.}}
    \label{fig:50percent}
\end{figure}

Figure~\ref{fig:50percent} compares the average time to reach a coverage of $50\%$ for various values of $\alpha$ for $N=20$ robots. Note that the average time increases when $\alpha$ increases. \textcolor{black}{We note that as $\alpha \to 2^-$ we recover the ordinary Laplacian relevant for Brownian strategies \cite{stinga2019}.} The macroscopic simulation falls within one standard deviation from the Webots individual agent simulations, justifying the statistical significance of the results.\\

\begin{figure}
    \centering
    \includegraphics[width = 0.75\textwidth]{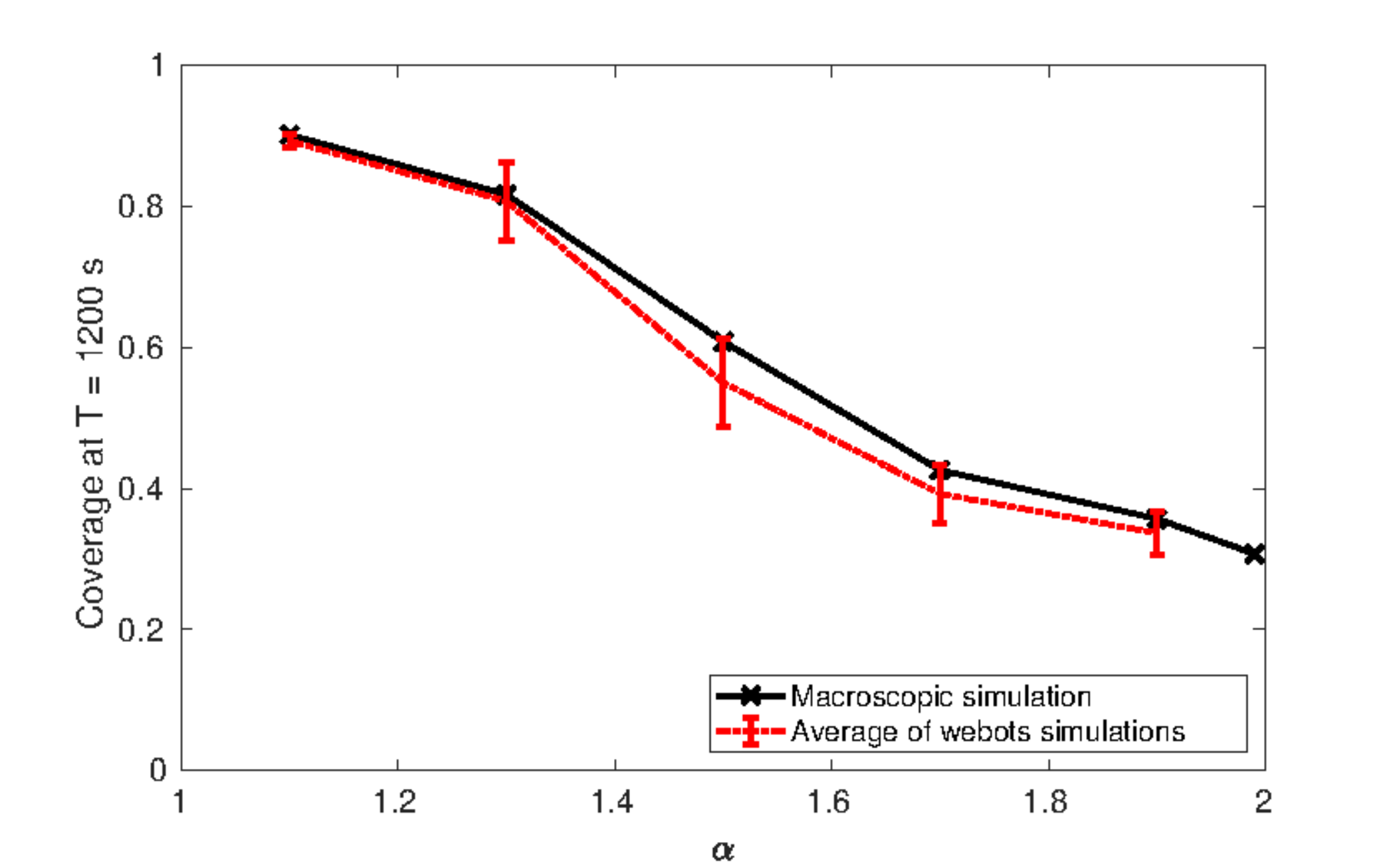}
    \caption{Comparison of coverage efficiency as a function of $\alpha$ for $N=20$ robots at $T = 1200 s$.  {The error bars show one standard deviation of the Webots simulations.}}
    \label{fig:cov_vs_alpha_N20}
\end{figure}

\begin{figure}
    \centering
    \includegraphics[width = 0.75\textwidth]{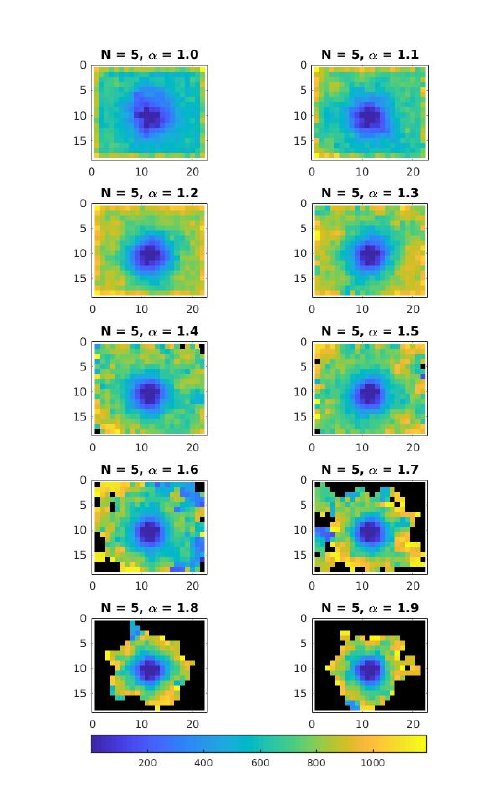}
    \caption{Expected hitting times taken from the Webots simulation at $T=1200 s$ for $N=5$ and different $\alpha$. Yellow colour corresponds to longer hitting times, blue corresponds to shorter hitting times\textcolor{black}{, and black corresponds to tiles which have not been reached.}}
    \label{fig:CovPerSq}
\end{figure}

In Figure~\ref{fig:cov_vs_alpha_N20} we compare the coverage of the Webots simulations with $N=20$ against the continuous model for different values of $\alpha$ at the final time $T=1200 s$. Note that the continuous model falls within one standard deviation from the individual robot simulation and exhibits a decreasing trend of coverage as a function of $\alpha$.\\
Figure~\ref{fig:CovPerSq} illustrates the hitting time efficiency for different values of $\alpha$ at $T=1200 s$ for $N=5$ robots based on the individual {robot} simulations. We note that as the value of $\alpha$ increases the hitting time of each tile increases. In particular, for values of $\alpha$ close to $2$, a significant part of the arena remains to be covered, as shown in black. As the value of $\alpha$ approaches $1$ large expected hitting times (in yellow) are only observed close to the boundary of the domain. This again underlines the advantage of long-range L\'{e}vy strategies for efficient coverage.\\

\begin{figure}
    \centering
    \includegraphics[width = 0.75\textwidth]{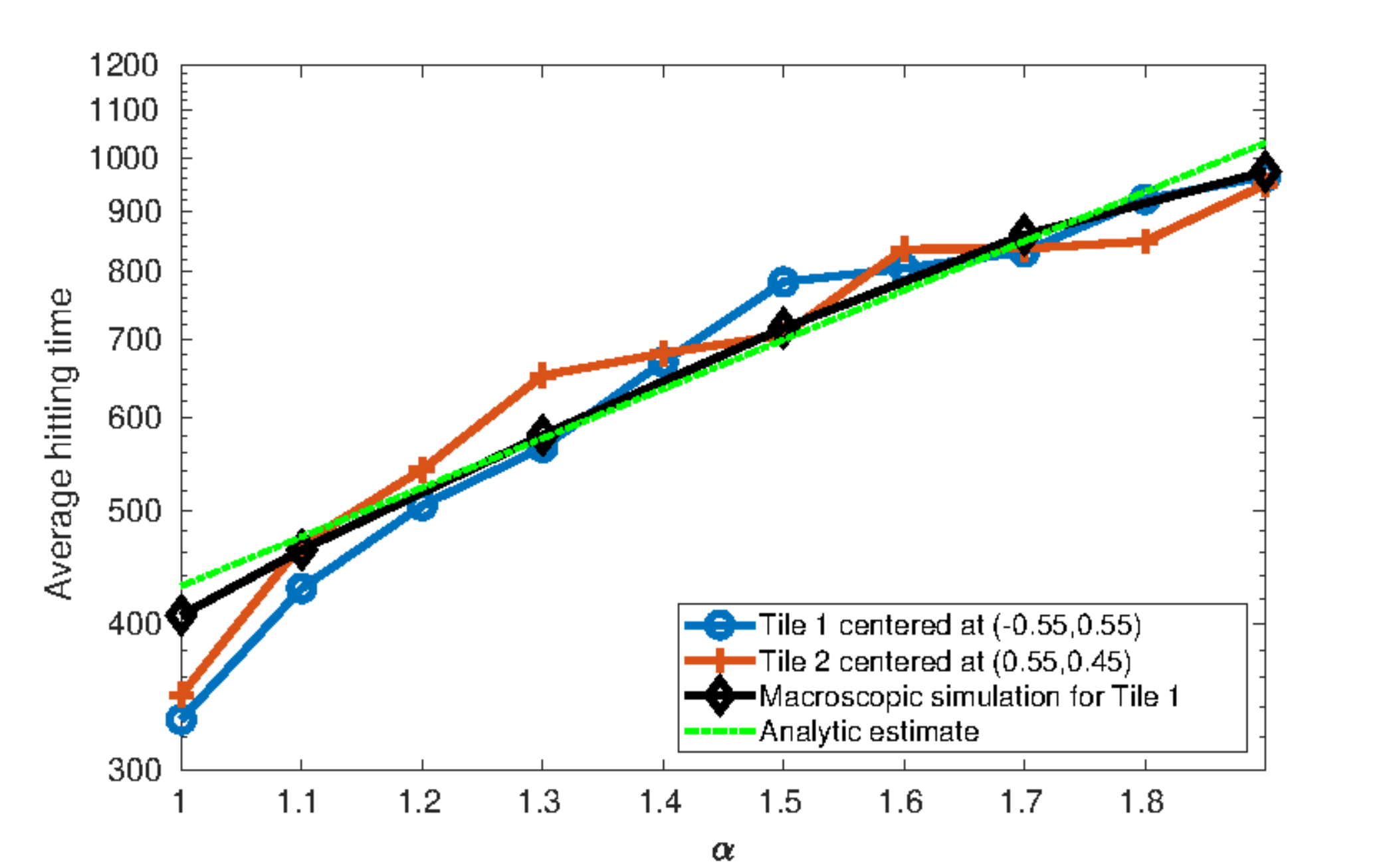}
    \caption{Hitting times with $N=5$ robots for two tiles in the domain centered at $(-0.55, 0.55)$ and $(0.55, 0.45)$, compared to the analytic solution in \eqref{eq:hit_time}. } 
    \label{fig:Hitting times}
\end{figure}

Figure~\ref{fig:Hitting times} shows the increase in the average hitting time for different values of $\alpha$ for two fixed tiles sized $10 \times 10 cm$ in the arena, centered at $(-0.55, 0.55)$, respectively $(0.55, 0.45)$ as highlighted in Figure~\ref{fig:domain}. Curves for tile 1 and tile 2 are generated from average of 70 runs for each value of $\alpha$. The Webots simulations are compared to the macroscopic model as well as the analytic approximation of the hitting times given by the following explicit formula obtained in \cite{m3as}:
\begin{equation}\label{eq:hit_time}
    t _ { 0 } \simeq \frac { \delta \pi } { 2 ^ { \alpha } \hat{ C } _ { \alpha} \operatorname { vol } ( \mathsf{T} ) \sum _ { i } \left| \mathbf { x } _ { 0 } - \mathbf { x } _ { i } \right| ^ { - \alpha - 2 } } \ .
\end{equation}
Here $\mathbf{x}_0$ is the centre of the target $\mathsf{T}$, $\mathbf{x}_i$ corresponds to the initial positions of the robots and $$\hat{C}_\alpha=-2\sqrt{\pi}\cos\left(\frac{\pi\alpha}{2} \right)\frac{\Gamma\left(\frac{\alpha+1}{2} \right)}{\Gamma\left(\frac{\alpha+2}{2} \right)} C_\alpha\ .$$ The target $\mathsf{T}$ is considered to be covered when the solution reaches a prescribed threshold $\delta\in(0,1)$. \\
Note that the analytic approximation closely matches both the macroscopic and the individual agent simulations.

The favorable long-range behaviour of L\'{e}vy over Brownian strategies is shown by the increase in coverage for smaller values of $\alpha$. This is illustrated by the clear quantitative agreement of macroscopic modeling and Webots individual particle simulations which falls within one standard deviation of the statistical uncertainty.\\

\begin{figure}
    \centering
    \includegraphics[width = 0.5\textwidth]{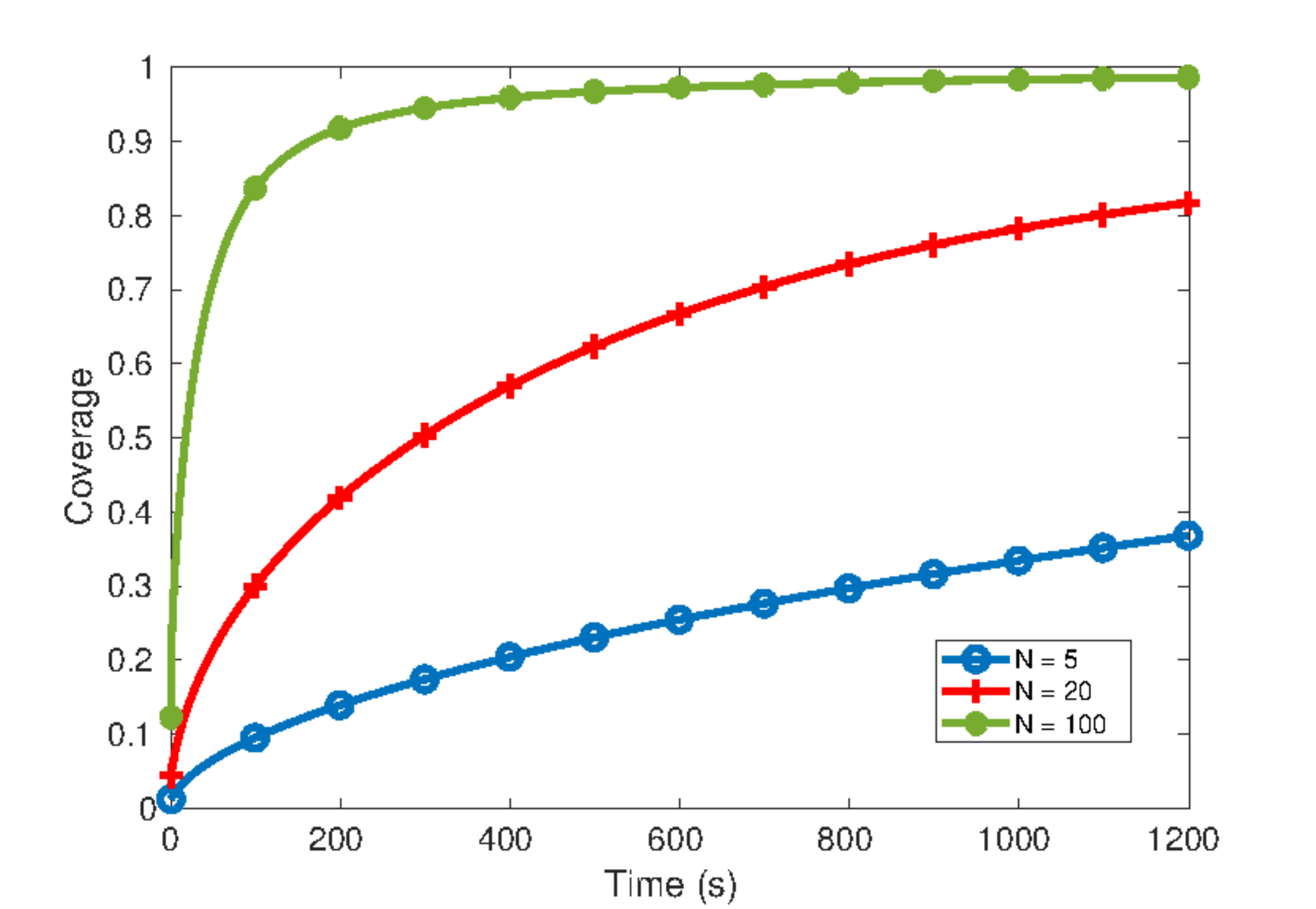}
    \caption{Comparison of macroscopic simulations for $N = 5, 20, 100$ and $\alpha = 1.3$.}
    \label{fig:Different N}
\end{figure}

In all simulations, the macroscopic modeling is faster and more efficient by orders of magnitude, \textcolor{black}{depending on the implementation and the hardware. To be specific, while the individual based particle simulations with Webots take a fixed fraction of the real time, depending on the number of robots $N$, the presented macroscopic simulations run within seconds, independent of $N$. Our} numerical results confirm the proposed definitions of coverage and hitting times. The efficient computation of these quantities is crucial for real-time optimization of robot movement strategies.\\
\textcolor{black}{The accuracy for the small systems previously considered tested the applicability and limitations of the macroscopic simulations in their most challenging regime. The macroscopic simulations then offer insight into large robotic systems at a fraction of the expense of Webots simulations. Comparison of three macroscopic simulations is presented in Figure~\ref{fig:Different N} for $\alpha = 1.3$ and $N = 5, 20, 100$ robots.}

We note that statistical deviations are larger for {smaller} numbers of robots, where also larger finite size effects are expected to play a role for the macroscopic models as can be seen in Figure~\ref{fig:comp_a13_Nvaried}. During intermediate times, statistical deviations are observed to be larger for small L\'{e}vy parameters $\alpha$. On the other hand, statistical certainty increases for longer times for values near full coverage. This is illustrated in Figure~\ref{fig:cov_vs_alpha_N20} for which the error bars are largest for intermediate values of $\alpha$. \\

\section{Discussion of results}

The simulation results in Section \ref{sec: comparison} show that Theorem \ref{thm: parabolic limit} leads to efficient computations of \textcolor{black}{quantities} like coverage and hitting times. Note, in particular, that the computational complexity of numerically approximating the differential equation in Theorem \ref{thm: parabolic limit} is independent of the number $N$ of robots. In fact, the  modeling accuracy increases with $N$, as finite size effects become less important. 

The speed up provided by the proposed approach, as compared to the Webots simulations, allows to assess strategies for robot movement and interactions. The formulation can be combined with standard optimization solvers  \cite{Troltzsch} to devise optimal strategies for a given task, as will be explored in forthcoming work.

Beyond the numerical assessment of specific strategies,  Theorem \ref{thm: parabolic limit} allows to use techniques from partial differential equations for the qualitative analysis of different movements strategies. This is exemplified by Equation \eqref{eq:hit_time} and Figure \ref{fig:Hitting times}, which give an explicit formula for the expected hitting time, in quantitative agreement with simulations. \textcolor{black}{Both the equation and the figure show} that the L\'{e}vy parameter should be chosen as small as possible, and how the hitting time depends on the geometry of the domain and the initial distribution of robots. \textcolor{black}{The simple relation obtained from the analysis is an example of the quantitative predictions possible for the macroscopic system which are not obvious from the movement laws. Further,} specific robot interactions can be analytically assessed based on whether and how they affect the resulting differential equation. For example, for given interaction and movement laws one may assess the optimal parameter ranges  for swarm formation,  or impossibility of it, without extensive simulation.

\textcolor{black}{The question of the optimal L\'{e}vy exponents for a given task and whether natural selection leads biological organisms to follow the optimal strategy has been of significant interest in different areas. A L\'{e}vy strategy with L\'{e}vy  exponent $\alpha=1$ was found optimal in \cite{viswanathan1999} for searching sparsely and randomly distributed revisitable targets in $2d$. While exponents $\alpha\leq1$ are not covered by the assumptions of current analysis, our findings are consistent with such previous results. A thorough use of the methods in this paper to find optimal L\'{e}vy exponents and their extension to L\'{e}vy exponents $\alpha\leq1$ remains for future work.}

\section{Conclusions}

In this paper we 
\textcolor{black}{apply} analysis tools developed for biological systems, like macroscopic modeling, \textcolor{black}{to} the assessment and design of swarm robotic systems. We illustrate our approach on automated search and area coverage applications. Starting from the behaviour of the individual robot, these tools allow the fast quantitative prediction of system-level properties and thereby provide a way towards the optimal engineering of a system for a given task, including qualitative analytical methods.

From the movement and interaction strategies of the individual robots we derive an effective diffusion equation for the density of robots on macroscopic length and time scales. We do so for a large class of L\'{e}vy strategies with characteristic long-range movement and compare it to regular diffusive strategies. The effective description allows the fast and accurate computation of quantities of robotic interest: We show that coverage and expected hitting times agree with the results of standard individual robot simulations within the statistical error margins of the latter, at \textcolor{black}{a fraction of the computational cost and with a run time independent of the number of robots}. 
{Compared to regular diffusive strategies, the characteristic long range movement in L\'{e}vy strategies leads to faster exploration of the spatial domain.} The results underline the exponential increase of the hitting time with the L\'{e}vy exponent, for constant robot speed, as confirmed by matching analytical and numerical results. \textcolor{black}{In the considered examples the L\'{e}vy exponent $\alpha =1$ leads to optimal area coverage and hitting times.}

{The fast simulation tools \textcolor{black}{used} in this article allow the efficient design of optimal strategies for a given problem and metric of success, see \cite{hinze}. Discretization of the continuous strategy then translates into a strategy for the individual robots, as explored in \cite{elamvazhuthi2018pde,zhang2018performance} and in ongoing work. 


\textcolor{black}{The presented macroscopic approach provides a way to overcome the computational limitations of particle simulations in the design of certain robotic swarms. We have shown how such an approach can be applied to search and coverage task for simple environments and simple, biologically-inspired controllers, since this may provide the basis for many of the applications envisaged for robotic swarms. Our validation by individual robot simulations confirms the applicability of our modeling for the robotic strategies under consideration, also for other tasks, metrics of success or in more complex domains \cite{estrada2018metaplex}. In general, it is far from straightforward to derive a set of equations for arbitrary control strategies and more complex settings \cite{bellomo2017active,carrillo2010particle}, and a validation would be required for newly modeled classes of robot interactions. The modeling effort should be weighted against the insights that could obtained with more straightforward computer simulations.} \textcolor{black}{The mathematical analysis of some specific interactions and strategies related to this article, such as long-range alignment and coordination, is considered in the article \cite{estrada2018swarming}. }

More generally, macroscopic models by partial differential equations have become a standard modeling tool in biological contexts \cite{codling2008random,hillen2009user}. Future work aims to adapt the available tool box towards other swarm robotics applications, including pheromone cues, long-range coordination \cite{estrada2018swarming}, \textcolor{black}{strategies for complex domains, such as exploration of unknown and hazardous environments \cite{estrada2018metaplex},} and validation with real robot systems. 

Work in \cite{andersonquantitative,brambilla} has started to investigate  metrics relevant to robotics applications. Also direct approaches to the control of the interacting particle system are a topic of current interest \cite{bailo2018optimal}.}







\ack
The authors would like to thank J.~A.~Carrillo, K.~J.~Painter and H.~Sardinha for fruitful discussions and H.~Fraser and M.~A.~Campbell for technical assistance. \\

\section*{References}
\begin{thebibliography}{99}

\bibitem{Acosta1}
Gabriel Acosta and Juan~Pablo Borthagaray.
\newblock A fractional {L}aplace equation: regularity of solutions and finite
  element approximations.
\newblock {\em SIAM Journal on Numerical Analysis}, 55(2):472--495, 2017.

\bibitem{Acosta2}
G. Acosta, J.~P. Borthagaray, N. Heuer.\
\newblock {Finite element approximations of the nonhomogeneous fractional Dirichlet problem}
\newblock {\em IMA Journal of Numerical Analysis}, 39(3):1471--1501, 2019.



\bibitem{alt1980biased}
Wolfgang Alt.
\newblock Biased random walk models for chemotaxis and related diffusion
  approximations.
\newblock {\em Journal of Mathematical Biology}, 9(2):147--177, 1980.


\bibitem{altshuler2005swarm}
Yaniv Altshuler, Alfred~M Bruckstein, and Israel~A Wagner.
\newblock Swarm robotics for a dynamic cleaning problem.
\newblock In {\em Swarm intelligence in intrusion detection: A survey, 2011.
  Computers \& Security,}, pages 209--216. IEEE, 2005.



\bibitem{andersonquantitative}
Brendon~G Anderson, Eva Loeser, Marissa Gee, Fei Ren, Swagata Biswas, Olga
  Turanova, Matt Haberland, and Andrea~L Bertozzi.
\newblock Quantitative assessment of robotic swarm coverage.
\newblock {\em arXiv preprint, arXiv:1806.02488}, 2018.

\bibitem{bailo2018optimal}
Rafael Bailo, Mattia Bongini, Jos{\'e} A. Carrillo,  and Dante Kalise.
\newblock Optimal consensus control of the Cucker-Smale model.
\newblock {\em arXiv preprint, arXiv:1802.01529}, 2018.


\bibitem{brambilla}
Manuele Brambilla, Eliseo Ferrante, Mauro Birattari, and Marco Dorigo.
\newblock Swarm robotics: a review from the swarm engineering perspective.
\newblock {\em Swarm Intelligence}, 7:1--41, 2013.

\bibitem{cercignani2013mathematical}
Carlo Cercignani, Reinhard Illner, and Mario Pulvirenti.
\newblock {\em The mathematical theory of dilute gases}, volume 106.
\newblock Springer Science \& Business Media, 2013.

\bibitem{chambers1976method}
Chambers, John M and Mallows, Colin L and Stuck, B W
\newblock{\em A method for simulating stable random variables}, volume 71, No 354, 340--344
\newblock Journal of the American Statistical Association,1976


\bibitem{chuang2007state}
Yao-li Chuang,  Maria R. D’orsogna,  Daniel Marthaler,  Andrea L. Bertozzi  and Lincoln S. Chayes. 
\newblock State transitions and the continuum limit for a 2D interacting, self-propelled particle system.
\newblock {\em Physica D: Nonlinear Phenomena}, 232(1):33--47, 2007.


\bibitem{codling2008random}
Edward A. Codling, Michael J. Plank and Simon Benhamou.
\newblock Random walk models in biology.
\newblock {\em Journal of the Royal Society Interface}, 5(25):813--834, 2008.





\bibitem{couzin2002collective}
Iain D. Couzin,  Jens Krause, Richard James,  Graeme D. Ruxton  and Nigel R. Franks.
\newblock Collective memory and spatial sorting in animal groups.
\newblock {\em Journal of Theoretical Biology}, 218(1):1--11, 2002.

\bibitem{cucker2007emergent}
Felipe Cucker and Steve Smale. 
\newblock Emergent behavior in flocks.
\newblock {\em IEEE Transactions on automatic control}, 52(5):852--862, 2007.




\bibitem{carrillo2010particle}
Jos{\'e} A Carrillo, Massimo Fornasier,  Giuseppe Toscani and Francesco  Vecil.
\newblock Particle, kinetic, and hydrodynamic models of swarming.
\newblock {\em Mathematical modeling of collective behavior in socio-economic and life sciences}, Springer, 297--336, 2010.



\bibitem{degond2008continuum}
Pierre Degond and S{\'e}bastien  Motsch.
\newblock Continuum limit of self-driven particles with orientation interaction.
\newblock {\em Mathematical Models and Methods in Applied Sciences}, 18(supp01):1193--1215, 2008.

\bibitem{dhariwal2004bacterium}
Amit Dhariwal, Gaurav~S Sukhatme, and Aristides~AG Requicha.
\newblock Bacterium-inspired robots for environmental monitoring.
\newblock In {\em Robotics and Automation, 2004. Proceedings. ICRA'04. 2004
  IEEE International Conference}, 1436--1443. IEEE, 2004.


\bibitem{dorigo2014swarm}
Marco Dorigo, Mauro Birattari, and Manuele Brambilla.
\newblock Swarm robotics.
\newblock {\em Scholarpedia}, 9(1):1463, 2014.


\bibitem{elamvazhuthi2016coverage}
K.~Elamvazhuthi, C.~Adams, and S.~Berman, ``Coverage and field estimation on bounded domains by diffusive swarms,''
  \emph{2016 IEEE 55th Conference on Decision and Control (CDC)}, vol.~95, pp. 2867--2874, 2016.

\bibitem{elamvazhuthi2018pde}
K.~Elamvazhuthi, H.~Kuiper, and S.~Berman, ``PDE-based optimization for
  stochastic mapping and coverage strategies using robotic ensembles,''
  \emph{Automatica}, vol.~95, pp. 356--367, 2018.

\bibitem{elamvazhuthi2018nonlinear}
K.~Elamvazhuthi and S.~Berman, ``Nonlinear generalizations of diffusion-based
  coverage by robotic swarms,'' in \emph{2018 IEEE Conference on Decision and
  Control (CDC)}.\hskip 1em plus 0.5em minus 0.4em\relax IEEE, 2018, pp.
  1341--1346.

\bibitem{fossum2014repellent}
Filip Fossum, Jean-Marc Montanier,  and Pauline C. Haddow.
\newblock Repellent pheromones for effective swarm robot search in unknown environments.
\newblock In {\em 2014 IEEE Symposium on Swarm Intelligence (SIS)} 1-8. IEEE, 2014 

\bibitem{fioriti2015levy}
Fioriti, Vincenzo and Fratichini, Fabio and Chiesa, Stefano and Moriconi, Claudio
\newblock {\em Levy foraging in a dynamic environment--extending the Levy search}
\newblock {\em International Journal of Advanced Robotic Systems}, 12, No 7 pp 98, 2015.



\bibitem{fredslund2001robot}
Jakob Fredslund and Maja~J Mataric.
\newblock Robot formations using only local sensing and control.
\newblock In {\em Computational Intelligence in Robotics and Automation, 2001.
  Proceedings 2001 IEEE International Symposium on}, pages 308--313. IEEE,
  2001.
  
  
\bibitem{ha2008particle}
Seung-Yeal Ha  and Eitan Tadmor. 
\newblock From particle to kinetic and hydrodynamic descriptions of flocking.
\newblock {\em Kinetic \& Related Models}, 1(3):415--435, 2008.
 

\bibitem{hillen2009user}
Thomas Hillen and Kevin J. Painter. 
\newblock A user’s guide to PDE models for chemotaxis.
\newblock {\em Journal of Mathematical Biology}, 58(1--2):183, 2009.

\bibitem{kantor2003distributed}
George Kantor, Sanjiv Singh, Ronald Peterson, Daniela Rus, Aveek Das, Vijay
  Kumar, Guilherme Pereira, and John Spletzer.
\newblock Distributed search and rescue with robot and sensor teams.
\newblock In {\em Field and Service Robotics}, pages 529--538. Springer, 2003.


\bibitem{kolias2011swarm}
Constantinos Kolias, Georgios Kambourakis, and M~Maragoudakis.
\newblock Swarm intelligence in intrusion detection: A survey.
\newblock {\em Computers \& Security}, 30(8):625--642, 2011.


\bibitem{DBLP:journals/corr/KrivonosovDZ16}
M. Krivonosov, S. Denisov, V. Zaburdaev.
\newblock L{\'{e}}vy robotics.
\newblock {\em arXiv}, 1612.03997, 2016.

\bibitem{liu2007modeling}
Wenguo Liu, Alan~FT Winfield, and Jin Sa.
\newblock Modelling swarm robotic systems: A case study in collective foraging.
\newblock {\em Towards Autonomous Robotic Systems}, pages 25--32, 2007.

  
\bibitem{Elamvazhuthi_2019}
K.~Elamvazhuthi and S.~Berman.
\newblock Mean-field models in swarm robotics: a survey.
\newblock Bioinspiration {\&} Biomimetics 15: 015001, 2019.

\bibitem{francesca2016automatic}
Gianpiero Francesca and  Mauro Birattari
\newblock Automatic design of robot swarms: achievements and challenges.
\newblock Frontiers in Robotics and AI 3, 2016.

  

\bibitem{estrada2018metaplex}
Gissell Estrada-Rodriguez, Ernesto Estrada and Heiko Gimperlein.
\newblock Metaplex networks: Influence of the exo-endo structure of complex systems on diffusion.
\newblock {\em SIAM Review}, to appear, 2020.


\bibitem{estrada2018swarming}
Gissell Estrada-Rodriguez and Heiko Gimperlein.
\newblock Swarming of interacting robots with L\'{e}vy strategies: a
  macroscopic description.
\newblock {\em SIAM Journal on Applied Mathematics} 80, pages 476--498, 2020.

\bibitem{m3as}
Gissell Estrada-Rodriguez, Heiko Gimperlein, Kevin~J.~Painter, and Jakub Stocek.
\newblock Space-time fractional diffusion in cell movement models with delay.
\newblock {\em Mathematical Models and Methods in Applied Sciences} 29, pages 65--88, 2019.


\bibitem{franz2016hard}
Benjamin Franz, Jake~P Taylor-King, Christian Yates, and Radek Erban.
\newblock Hard-sphere interactions in velocity-jump models.
\newblock {\em Physical Review E}, 94(1):012129, 2016.

\bibitem{fricke2016immune}
George~Matthew Fricke, Joshua~P. Hecker, Judy~L. Cannon, and Melanie~E. Moses.
\newblock Immune-inspired search strategies for robot swarms.
\newblock {\em Robotica}, 34(8):1791--1810, 2016.

\bibitem{VIpreprint}
Heiko Gimperlein and Jakub Stocek.
\newblock Space-time adaptive finite elements for nonlocal parabolic
  variational inequalities.
\newblock {\em Computer Methods in Applied Mechanics and Engineering} 352, pages 137--171, 2019.

\bibitem{Hamann2006analytical}
H.~Hamann and H.~W{\"o}rn, ``An analytical and spatial model of foraging in a
  swarm of robots,'' in \emph{International workshop on swarm robotics}.\hskip
  1em plus 0.5em minus 0.4em\relax Springer, 2006, pp. 43--55.



\bibitem{harris}
Tajie Harris et~al.
\newblock Generalized L\'{e}vy walks and the role of chemokines in migration of effector CD8+ T cells.
\newblock {\em Nature}, pages 545--548, volume 486, 2012.

\bibitem{hinze}
M. Hinze, R. Pinnau, M. Ulbrich and S. Ulbrich
\newblock {\em Optimization with PDE Constraints}
\newblock Springer, 2009.


\bibitem{marthaler2004levy}
Marthaler, Daniel and Bertozzi, Andrea L and Schwartz, Ira B
\newblock{\em Levy searches based on a priori information: The biased Levy walk}.
\newblock{CALIFORNIA UNIV LOS ANGELES DEPT OF MATHEMATICS}, 2004

\bibitem{marjovi2009multi}
Ali Marjovi, Jo{\~a}o~Gon{\c{c}}alo Nunes, Lino Marques, and An{\'\i}bal
  de~Almeida.
\newblock Multi-robot exploration and fire searching.
\newblock In {\em Intelligent Robots and Systems, 2009. IROS 2009. IEEE/RSJ
  International Conference on}, pages 1929--1934. IEEE, 2009.

\bibitem{mataric1997reinforcement}
Maja~J Matari{\'c}.
\newblock Reinforcement learning in the multi-robot domain.
\newblock In {\em Robot colonies}, pages 73--83. Springer, 1997.


\bibitem{meerschaert2006fractional}
Mark M. Meerschaert, Jeff Mortensen and  Stephen W. Wheatcraft.
\newblock Fractional vector calculus for fractional advection--dispersion.
\newblock {\em Physica A: Statistical Mechanics and its Applications}, 367:181--190, 2006.

\bibitem{mes1} A. R. Mesquita, J. P. Hespanha and K. Astrom. 
\newblock Optimotaxis: a stochastic multi-agent optimization procedure with point measurements 
\newblock Int. Workshop on Hybrid Systems: Computation and Control (Springer), pages 358--371, 2008.

\bibitem{mes} A. R. Mesquita and J. P. Hespanha, 
\newblock Jump Control of Probability Densities With Applications to Autonomous Vehicle Motion.
\newblock IEEE Transactions on Automatic Control 57(10), pages 2588--2598,  2012.

\bibitem{webots}
Olivier Michel.
\newblock Cyberbotics {L}td. Webots™: professional mobile robot simulation.
\newblock {\em International Journal of Advanced Robotic Systems}, 1(1):5,
  2004.

\bibitem{milutinovi2006modeling}
D.~Milutinovi and P.~Lima, ``Modeling and optimal centralized control of a
  large-size robotic population,'' \emph{IEEE Transactions on Robotics},
  vol.~22, no.~6, pp. 1280--1285, 2006.

\bibitem{bellomo2017active}
Nicola Bellomo, Pierre Degond and Eitan Tadmor
\newblock Active Particles, Volume 1: Advances in Theory, Models, and Applications, \emph{Birkh{\"a}user}, 2017.
 
\bibitem{epuck}
F.~Mondada et~al.
\newblock The e-puck, a robot designed for education in engineering.
\newblock In {\em Proceedings of the 9th Conference on Autonomous Robot Systems
  and Competitions}, pages 59--65, 2009.
  
\bibitem{NIST:DLMF} 
NIST Digital Library of Mathematical Functions
\newblock http://dlmf.nist.gov/, Release 1.0.14 of 2016-12-21
\newblock F.~W.~J. Olver, A.~B. {Olde Daalhuis}, D.~W. Lozier, B.~I. Schneider,
               R.~F. Boisvert, C.~W. Clark, B.~R. Miller and B.~V. Saunders, eds.
  


\bibitem{nolfi2000evolutionary}
Stefano Nolfi and Dario Floreano.
\newblock {\em Evolutionary robotics: The biology, intelligence, and technology
  of self-organizing machines}.
\newblock MIT press, 2000.


\bibitem{lerman2004review}
 Kristina Lerman, Alcherio Martinoli and  Aram Galstyan.
\newblock A review of probabilistic macroscopic models for swarm robotic systems
\newblock {\em International Workshop on Swarm Robotics}, pages 143--152, 2004.


\bibitem{schroeder2017efficient}
Adam Schroeder, Subramanian Ramakrishnan, Manish Kumar and Brian Trease.
\newblock Efficient spatial coverage by a robot swarm based on an ant foraging model and the L{\'e}vy distribution
\newblock {\em Swarm Intelligence}, No. 11, pages 39--69, 2017.


\bibitem{nurzaman2009yuragi}
  Surya G. Nurzaman,  Yoshio Matsumoto,  Yutaka Nakamura,  Satoshi Koizumi  and Hiroshi Ishiguro.
\newblock Yuragi-based adaptive searching behavior in mobile robot: From bacterial chemotaxis to Levy walk.
\newblock {\em Robotics and Biomimetics, 2008. ROBIO 2008. IEEE International Conference on}, 806--811, 2009.

\bibitem{othmer2000diffusion}
Hans~G Othmer and Thomas Hillen.
\newblock The diffusion limit of transport equations derived from velocity-jump
  processes.
\newblock {\em SIAM Journal on Applied Mathematics}, 61(3):751--775, 2000.




\bibitem{parker1995design}
Lynne~E Parker.
\newblock On the design of behavior-based multi-robot teams.
\newblock {\em Advanced Robotics}, 10(6):547--578, 1995.

\bibitem{parker2008distributed}
Lynne~E Parker.
\newblock Distributed intelligence: Overview of the field and its application
  in multi-robot systems.
\newblock {\em Journal of Physical Agents}, 2(1):5--14, 2008.

\bibitem{Argos}
Carlo~Pinciroli, et al.
\newblock ARGoS: a Modular, Parallel, Multi-Engine Simulator for Multi-Robot Systems. 
\newblock {\em Swarm Intelligence}, 6 (4): 271--295, 2012.


\bibitem{prorok2011multi}
A.~Prorok, N.~Correll, and A.~Martinoli, ``Multi-level spatial modeling for
  stochastic distributed robotic systems,'' \emph{International Journal of
  Robotics Research}, vol.~30, no.~5, pp. 574--589, 2011.


\bibitem{schmickl2011cocoro}
Thomas Schmickl, Ronald Thenius, Christoph Moslinger, Jon Timmis, Andy Tyrrell,
  Mark Read, James Hilder, Jose Halloy, Alexandre Campo, Cesare Stefanini,
  et~al.
\newblock Cocoro--the self-aware underwater swarm.
\newblock In {\em Self-Adaptive and Self-Organizing Systems Workshops (SASOW),
  2011 Fifth IEEE Conference on}, pages 120--126. IEEE, 2011.



\bibitem{senanayake2016search}
Madhubhashi Senanayake, Ilankaikone Senthooran, Jan~Carlo Barca, Hoam Chung,
  Joarder Kamruzzaman, and Manzur Murshed.
\newblock Search and tracking algorithms for swarms of robots: {A} survey.
\newblock {\em Robotics and Autonomous Systems}, 75:422--434, 2016.

\bibitem{soysal2006macroscopic}
O.~Soysal and E.~{\c{S}}ahin, ``A macroscopic model for self-organized
  aggregation in swarm robotic systems,'' in \emph{International Workshop on
  Swarm Robotics}.\hskip 1em plus 0.5em minus 0.4em\relax Springer, 2006, pp.
  27--42.

\bibitem{stinga2019}  
P.~R.~Stinga.
\newblock {User’s Guide to the Fractional Laplacian and the Method of Semigroups.} \newblock in {\em Fractional Differential Equations}, 235–266, 2019.

\bibitem{sutantyo2013collective}
   Donny Sutantyo, Paul Levi, Christoph M{\"o}slinger  and Mark Read.
\newblock Collective-adaptive L{\'e}vy flight for underwater multi-robot exploration.
\newblock {\em 2013 IEEE International Conference on Mechatronics and Automation (ICMA)}, 456--462, 2013.



\bibitem{Troltzsch}
Fredi Tr\"{o}ltzsch.
\newblock Optimal Control of Partial Differential Equations: Theory, Methods, and Applications.
\newblock {\em American  Mathematical  Society}, 112, 2010.

\bibitem{turduev2010chemical}
Mirbek Turduev, Murat Kirtay, Pedro Sousa, Veysel Gazi, and Lino Marques.
\newblock Chemical concentration map building through bacterial foraging
  optimization based search algorithm by mobile robots.
\newblock In: {\em Systems Man and Cybernetics (SMC), 2010 IEEE International
  Conference}, pages 3242--3249. IEEE, 2010.

\bibitem{vargas2014her}
Patricia A. Vargas, Ezequiel A. Di Paolo,  Inman Harvey and Phil Husbands.
\newblock The Horizons of Evolutionary Robotics.
\newblock {\em The MIT Press}, 1st, 2015.
\bibitem{vicsek1995novel}
 Tam{\'a}s Vicsek, Andr{\'a}s Czir{\'o}k, Eshel Ben-Jacob, Inon Cohen  and Ofer Shochet.
\newblock Novel type of phase transition in a system of self-driven particles.
\newblock {\em Physical Review Letters}, 75(6):1226, 1995.

\bibitem{viswanathan1999}
G. Viswanathan, S. Buldyrev, S. Havlin, et al.
\newblock {Optimizing the success of random searches.}
\newblock {\em Nature}, 401:911–914, 1999.


\bibitem{zhang2018performance}
Fangbo Zhang, Andrea~L Bertozzi, Karthik Elamvazhuthi, and Spring Berman.
\newblock Performance bounds on spatial coverage tasks by stochastic robotic
  swarms.
\newblock {\em IEEE Transactions on Automatic Control}, 63(6):1563--1578, 2018.

\bibitem{epuckedu82:online}
GCTronic.
\newblock e-puck education robot - an overview.
\newblock \url{http://www.e-puck.org/}.
\newblock (Accessed on 03/01/2019).

\bibitem{IRProxim47:online}
GCTronic.
\newblock epuck robot with proximity sensors description.
\newblock \url{http://www.e-puck.org/index.php?option=com_content&view=article&id=22&Itemid=13}.
\newblock (Accessed on 03/01/2019).
https://www.overleaf.com/project/5e7bd0205c96690001cc4ef3

\end{thebibliography}
%

\appendix
\section{Turn angle operator}\label{sec: turn_angle_properties}
 
This section recalls some basic spectral properties of the turn angle operator $T$ defined in (\ref{eq: turn angle operator}). Crucially, its kernel $k(\mathbf{\theta};\mathbf{\eta}) =  \tilde{k}(|\eta-\theta|)$ only depends on the distance $|\eta-\theta|$: 
\begin{align*}
T\phi(\eta) & =\int_S k(\mathbf{\theta};\mathbf{\eta})\phi(\theta)d\theta = \int_S \tilde{k}(|\eta-\theta|)\phi(\theta)d\theta.
\end{align*}
Because $\tilde{k}$ is a probability distribution, it is normalized to $\int_S \tilde{k}(|\theta-e_1|)d\theta=1$,  where $e_1 = (1,0,\dots, 0)$. We immediately observe
\begin{align}
    \int_S (\mathds{1}-T)\phi d\theta=0\label{eq: conservation T}
\end{align}
for all $\phi\in L^2(S)$. Expression (\ref{eq: conservation T}) corresponds to the conservation of the number of robots after reorientations.

We also require some more detailed information about the spectrum of $T$. Recall that in $n$-dimensions, the surface area of the unit sphere $S$ is given by \[
|S|=\begin{cases}
\frac{2\pi^{\nicefrac{n}{2}}}{\Gamma\left(\frac{n}{2}\right)}, & \textnormal{for $n$ even},\\
\frac{\pi^{\nicefrac{n}{2}}}{\Gamma\left(\frac{n}{2}+1\right)}, & \textnormal{for $n$ odd}.
\end{cases}
\]

\section{Derivation of the one-particle transport equation}\label{sec: derivation}

Integrating with respect to $\mathbf{x}_2$ and $\theta_2$ in  (\ref{eq: system}), from (I) we can commute the integrals and the time derivative to obtain $|S|\partial_tp$. Using Reynolds' {transport theorem in the variable $\mathbf{x}_1$} in (II) and the divergence theorem in (III) we obtain
\begin{align}
c\int_{\Omega_2}&\int_S(\theta_1\cdot\nabla_{\mathbf{x}_1}\tilde{\tilde{\sigma}} +\theta_2\cdot\nabla_{\mathbf{x}_2}\tilde{\tilde{\sigma}})d\theta_2d\mathbf{x}_2 = |S|c\theta_1\cdot\nabla p\nonumber\\ &-c\int_{B_\varrho}\int_S(\theta_1\cdot\nu)\tilde{\tilde{\sigma}}d\theta_2d\mathbf{x}_2+c\int_{B_\varrho}\int_S(\theta_2\cdot\nu)\tilde{\tilde{\sigma}}d\theta_2d\mathbf{x}_2\ \label{eq: derivation 1}
\end{align}
where $\nu$ is the outward pointing unit normal vector with respect to $\Omega_2$. From this computation we obtain the first term in the right hand side of (\ref{eq: one particle density}).

Using the conservation of the number of robots after reorientations given by (\ref{eq: conservation T}), the term (V) in (\ref{eq: system}) vanishes after integration with respect to $\theta_2$.

After integration, the term (IV) can be written as

\[  
-(\mathds{1}-T_1)\int_0^t\beta_1(\mathbf{x}_1,\tau_1)i(\mathbf{x}_1,t,\theta_1,\tau_1)d\tau_1\ ,
\]
where $i(\mathbf{x}_1,t,\theta_1,\tau_1)=\int_0^t\int_{\Omega_2}\int_S\sigma d\theta_2d\mathbf{x}_2d\tau_2$. Following standard arguments from \cite{estrada2018swarming} we can write the above expression as a convolution in time
\begin{equation}
-(\mathds{1}-T_1)\int_0^t\mathcal{B}(\mathbf{x}_1,t-s)p(\mathbf{x}_1-c\theta_1(t-s),s,\theta_1)ds\label{eq: derivation 2}
\end{equation}
where the operator $\mathcal{B}$ is defined from its Laplace transform $\hat{\mathcal{B}}=\mathcal{L}\{\mathcal{B} \}$ in time,
\begin{equation}
    \hat{\mathcal{B}}(\mathbf{x}_1,\lambda+c\theta_1\cdot\nabla_{\mathbf{x}_1})=\frac{\hat{\varphi}_1(\mathbf{x}_1,\lambda+c\theta_1\cdot\nabla_{\mathbf{x}_1})}{\hat{\psi}_1(\mathbf{x}_1,\lambda+c\theta_1\cdot\nabla_{\mathbf{x}_1})}\ ,\label{eq: operator B}
\end{equation}
where $\lambda$ is the Laplace variable, $\psi_1$ is given in (\ref{eq: survival}) and $\varphi_1=-\partial_{\tau_1}\psi_1$. 

Combining (\ref{eq: derivation 1}) and (\ref{eq: derivation 2}) then equation (\ref{eq: one particle density}) is obtained from (\ref{eq: system}),  after introducing the scaling.

\section{Fractional operator: definition and derivation}\label{app: fractiona laplacian}
We recall some basic definitions concerning fractional differential operators.

\begin{definition}\label{def: fractional}
For $s \in (0,2)$ and $f \in C^2(\mathds{R}^n)$ define the fractional gradient of $f$ as
\begin{equation}
\nabla^s f(\mathbf{x})=\frac{1}{|S|}\int_{S}\mathbf{\theta}\mathbf{D}_{\mathbf{\theta}}^s f(\mathbf{x})d\mathbf{\theta}=\frac{1}{|S|}\int_{S}\mathbf{\theta}(\mathbf{\theta}\cdot\nabla)^s f(\mathbf{x})d\mathbf{\theta},\label{eq: fractional derivative}
\end{equation}
where $\mathbf{D}_{\mathbf{\theta}}^s=(\mathbf{\theta}\cdot\nabla)^s$ is the fractional directional derivative of order $s$.
The fractional Laplacian of $f$ is given by
\begin{equation}
\mathds{D}^s f(\mathbf{x})=\frac{1}{|S|}\int_{S}\mathbf{D}^s_\mathbf{\theta}f(\mathbf{x})d\mathbf{\theta}.\label{eq: fractional Laplacian}
\end{equation}
\end{definition}
$\mathds{D}^s$ is associated to $(-\Delta)^{\nicefrac{\alpha}{2}}$ in the following way,
$
\mathds{D}^sf(\mathbf{x})=\varXi_\alpha(-\Delta)^{\nicefrac{\alpha}{2}}
$
where, in two dimensions, for $1<\alpha<2$,
\begin{equation}
\varXi_\alpha=-2\sqrt{\pi}\cos\left(\frac{\pi\alpha}{2} \right)\frac{\Gamma\left(\frac{\alpha+1}{2} \right)}{\Gamma\left(\frac{\alpha+2}{2} \right)}.\label{appendixnormal}
\end{equation}
See \cite{meerschaert2006fractional} for further information.

The fractional Laplacian in (\ref{eq:final}) is obtained as follows. Using  the quasi-static approximation
\begin{equation}
\hat{\mathcal{B}}^\varepsilon(\mathbf{x}_1,\varepsilon\lambda+\varepsilon^{1-\gamma}c_0\theta_1\cdot\nabla)\simeq\hat{\mathcal{B}}^\varepsilon(\mathbf{x}_1,\varepsilon^{1-\gamma}c_0\theta_1\cdot\nabla)\ ,\label{eq: quasi-static app}
\end{equation}
we see that the convolution in time in the right hand side of (\ref{eq: one particle density}) vanishes.
We use (\ref{eq: operator B}) to write an explicit expression for (\ref{eq: quasi-static app}) based on the Laplace transforms of $\psi_1^\varepsilon$ and $\varphi_1^\varepsilon$ which are given by
\begin{equation*}
\hat{\psi}_1^\varepsilon(\mathbf{x}_1,\lambda)=a^\alpha \lambda^{\alpha-1} e^{a\lambda}\Gamma(-\alpha+1,a\lambda)\ \ \textnormal{and} \ \ \hat{\varphi}_1^\varepsilon(\mathbf{x}_1,\lambda) =\alpha (a\lambda)^\alpha\Gamma(-\alpha,a\lambda)e^{a\lambda}\ ,
\end{equation*}
{respectively, for $a=\varsigma_0\varepsilon^{\mu}$. Here we use an asymptotic expansion for the incomplete Gamma function  
\begin{eqnarray}
\Gamma(b,z) & = \Gamma(b)\left( 1-z^{b}e^{-z}\sum_{k=0}^{\infty}\frac{z^k}{\Gamma(b+k+1)}\right),\label{eq: 3.23}
\end{eqnarray}
for $b$ positive and not integer \cite{NIST:DLMF}.}

 We conclude
\begin{equation}
   \hat{\mathcal{B}}^\varepsilon(\mathbf{x}_1,\lambda)=\frac{\hat{\varphi}_1^\varepsilon(\mathbf{x}_1,\lambda)}{\hat{\psi}_1^\varepsilon(\mathbf{x}_1,\lambda)}=\frac{\alpha-1}{a}-\frac{\lambda}{2-\alpha}-a^{\alpha-2}\lambda^{\alpha-1}(\alpha-1)^2\Gamma(-\alpha+1)+\mathcal{O}(a^{\alpha-1}\lambda^\alpha)\ ,\label{eq: expansion T}
 \end{equation}
where $\lambda$ is in fact given by $\varepsilon^{1-\gamma}c_0\theta_1\cdot\nabla$. Finally, when we substitute the above expression into the right hand side of (\ref{eq: one particle density}), we multiply by $\theta_1$ and integrate over $S$ to find the mean direction $w(\mathbf{x}_1,t)$ (see Section \ref{sec: PDE model}) a term of the form $\int_S\theta_1(\theta\cdot\nabla)^{\alpha-1} u(\mathbf{x}_1,t)d\theta$  gives a  fractional gradient of the form defined in (\ref{eq: fractional derivative}).

\end{document}